\documentclass[10pt,twocolumn,letterpaper]{article}

\usepackage{cvpr}              

\usepackage{algorithm}
\usepackage{algpseudocode} 

\definecolor{cvprblue}{rgb}{0.21,0.49,0.74}
\usepackage[pagebackref,breaklinks,colorlinks,allcolors=cvprblue]{hyperref}

\def\paperID{20581} 
\def\confName{CVPR}
\def\confYear{2026}

\title{HandDreamer: Zero-Shot Text to 3D Hand Model Generation using Corrective Hand Shape Guidance}

\author{Green Rosh \quad  Prateek Kukreja$^{*}$\quad Vishakha SR$^{*}$ \quad Pawan Prasad B H \\
Samsung R\&D Institute India Bangalore\\
{\small Accepted to CVPR 2026}\\
}

\begin{document}
\twocolumn[{
\renewcommand\twocolumn[1][]{#1}%
\maketitle

\begin{center}
    \centering
    \captionsetup{type=figure}
    \includegraphics[width=\textwidth]{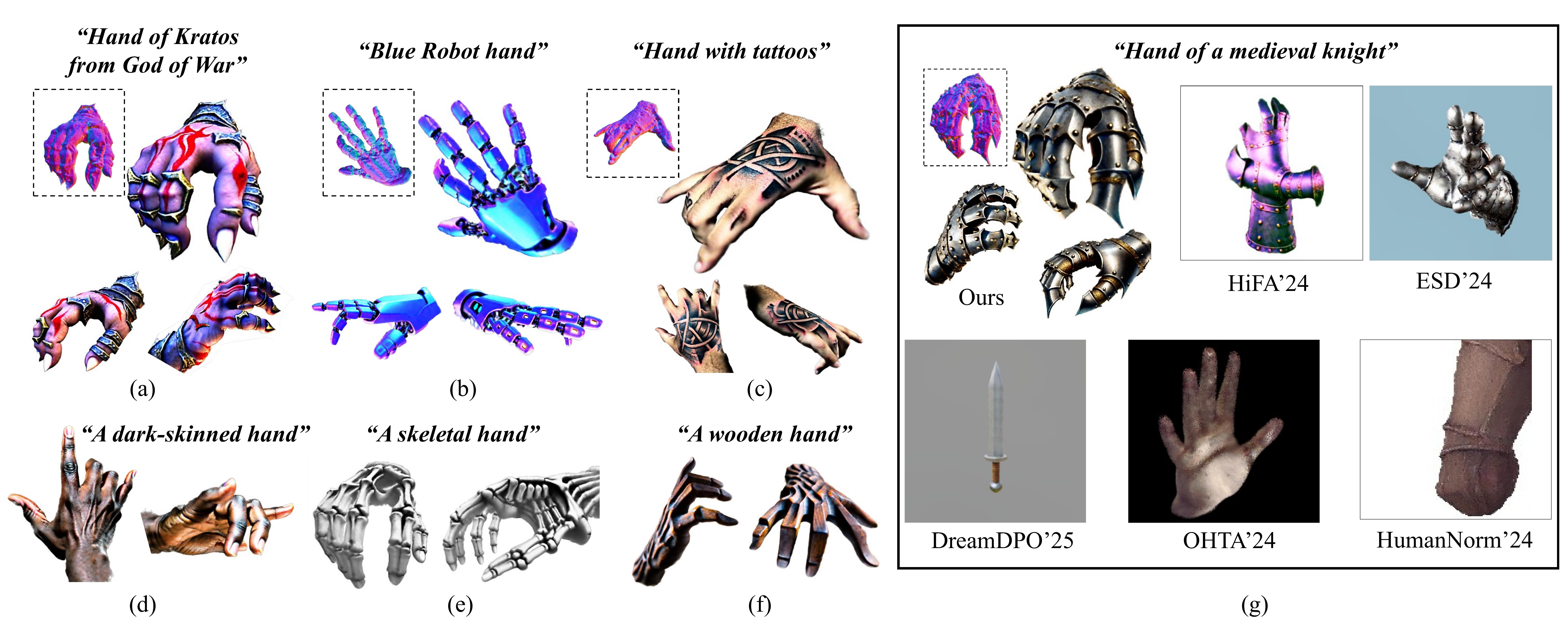}
    \captionof{figure}{We propose HandDreamer: the first method for zero-shot 3D hand generation from text prompts. Our method generates high-fidelity, geometrically accurate 3D hand models with diverse articulations from text prompts.  Existing methods generate Janus artifacts (HiFA, ESD) and fewer details (OHTA, DreamDPO, HumanNorm) (g). Surface maps provided inset.}
    \label{fig:poster}
\end{center}%
}]

\begingroup
\renewcommand\thefootnote{*}
\footnotetext{Equal Contribution}
\endgroup

\begin{abstract}
The emergence of virtual reality has necessitated the generation of detailed and customizable 3D hand models for interaction in the virtual world. However, the current methods for 3D hand model generation are both expensive and cumbersome, offering very little customizability to the users. While recent advancements in zero-shot text-to-3D synthesis have enabled the generation of diverse and customizable 3D models using Score Distillation Sampling (SDS), they do not generalize very well to 3D hand model generation, resulting in unnatural hand structures, view-inconsistencies and loss of details. To address these limitations, we introduce HandDreamer, the first method for zero-shot 3D hand model generation from text prompts. Our findings suggest that view-inconsistencies in SDS is primarily caused due to the ambiguity in the probability landscape described by the text prompt, resulting in similar views converging to different modes of the distribution. This is particularly aggravated for hands due to the large variations in articulations and poses. To alleviate this, we propose to use MANO hand model based initialization and a hand skeleton guided diffusion process to provide a strong prior for the hand structure and to ensure view and pose consistency.  Further, we propose a novel corrective hand shape guidance loss to ensure that all the views of the 3D hand model converges to view-consistent modes, without leading to geometric distortions. Extensive evaluations demonstrate the superiority of our method over the state-of-the-art methods, paving a new way forward in 3D hand model generation. 
\end{abstract}    
\section{Introduction}
\label{sec:intro}

With the advent of virtual reality and immersive realms, the world is moving towards a digital future. In this context, hands are indispensable for interactions in the virtual world where immersive telepresence and first-person video games are routine. Thus, there is a basic requirement for generating accurate 3D virtual assets representing hands, allowing users to customize their experience in the digital world. Conventional methods for creating such 3D hand models require multi-view capture systems with hundreds of cameras ~\cite{data1,data2}, along with expertise from graphics artists. This process is both extremely expensive and time consuming. Further, this makes it hard for users to create customized 3D hand models, thus limiting the democratization of 3D asset creation. To address these issues, there have been a focus on methods to generate 3D objects from text prompts using generative methods ~\cite{gan3D1,gan3D2,gan3D3} such as diffusion models ~\cite{diff1,diff2,diff3,stablediffusion,imagen}. However, the unavailability of large scale 3D datasets  hinder the progress of generative text-to-3D diffusion models.

To circumvent this challenge, a recent method, DreamFusion ~\cite{dreamfusion}, proposed Score Distillation Sampling (SDS) to leverage existing text-to-2D diffusion models to learn a 3D representation (eg: NeRF ~\cite{nerf}, Gaussian Splatting ~\cite{gausssplat}) of the required 3D scene. Several other methods propose enhancements ~\cite{fantasia,latentnerf,prolificdreamer,magic3D,dreamcraft} to improve the details of the generated 3D model. However, most of these methods suffer from viewpoint inconsistencies, resulting in multi-face ``Janus artifact" (Fig. \ref{fig:poster} (g-HiFA)). Other methods ~\cite{modeCollapse,dreamdpo} suggested joint-view approaches to minimize view-inconsistencies. However, we observed these approaches do not adapt well to highly articulatable objects such as hands, resulting in fingers protruding at unnatural locations (Fig. \ref{fig:poster} (g-ESD~\cite{modeCollapse})) and failing to generate hand semantics (Fig. \ref{fig:poster} (g-DreamDPO~\cite{dreamdpo})). There have also been 3D human model generation methods using SMPL ~\cite{smpl} based formulations. However, we observed that these methods generate hand regions with very less details (Fig. \ref{fig:poster} (g-HumanNorm)). There have also been methods ~\cite{harp,html,handy,ohta} which proposed to generate 3D hand models from one or more input images. However, we observed that these methods do not generalize well to handle fictional hands, resulting in stains and avatars that look like painted hands (Fig. \ref{fig:poster} (g-OHTA)). 

\begin{figure*}[tb]
  \centering
  \includegraphics[width=\textwidth]{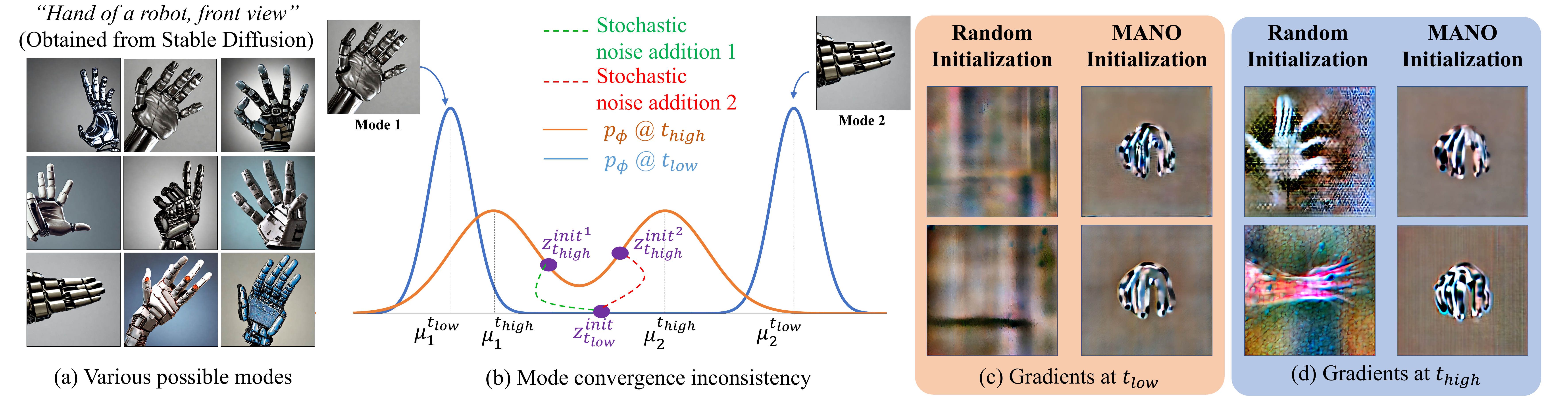}
  \caption{Convergence into wrong modes. (a) Probable modes for same view point. (b) Random initialization can converge into different modes for same viewpoints leading to Janus artifacts. (c,d) Visualization of gradients for the same viewpoint with multiple timesteps ($t$). Random initialization leads to less informative gradients at lower $t$ and diverse gradients leading to view-inconsistencies at higher $t$. MANO initialization yields consistent gradients at lower and higher $t$.}
  \label{fig:modeTheory}
\end{figure*}

In this paper, we first investigate the origin of viewpoint inconsistencies.  Our analysis suggests that the probability landscape defined by the text prompt consists of a large number of probable modes due to the huge variations in camera poses and hand articulations (Fig. \ref{fig:modeTheory}). We also observe that high-score initialization used in existing methods leads to similar views getting converged to different modes, resulting in view-inconsistencies ~\cite{hifa,dreamfusion}. We also observed that hand consists of severe self-occlusion, especially in side views, resulting in geometric degradation. To alleviate these challenges, we present HandDreamer, the first method to the best of our knowledge, for zero-shot 3D hand model generation. We first propose to condition the diffusion model using hand skeleton, embedding the camera and hand pose, so that the number of probable modes are reduced for each view point. Further, we also propose to create low-score initialization using a MANO hand model to ensure that all the views converge to the correct geometric mode. Finally, we also propose a novel Corrective Hand Shape (CHS) loss to ensure constant mode correction so that the geometry of the 3D model do not deviate too much from a plausible hand geometry. This loss helps alleviate geometric degradations especially in the side views of the hand. As shown in Fig. \ref{fig:poster}, the proposed HandDreamer method generates diverse 3D hand models with detailed texture and view-consistent geometry. We also show that our method generates superior results both quantitatively and qualitatively against state of the art text-to-3D methods. The generated 3D hand models can also be exported to hand meshes for downstream graphics tasks and articulations using tools such as Threestudio~\cite{threestudio}. The key contributions of this paper are as follows:
\begin{itemize}
  \item{We introduce HandDreamer, the first method to the best of our knowledge, for zero-shot 3D hand model generation from a text prompt and shape prior}
  \item{We provide an analysis of the origin of view-inconsistencies in text-to-3D hands and develop a novel corrective hand shape guided methodology to limit geometric degradations during SDS}
  \item{Our method generates high fidelity 3D hands outperforming existing methods qualitatively and quantitatively}
\end{itemize}

\section{Related Works}
\noindent\textbf{Text-to-3D content generation: } Earlier methods for text-to-3D generation ~\cite{clip1,clip2,clip3} proposed to utilize CLIP ~\cite{clip} model to learn the underlying 3D representation of the prompted scene. Recently, DreamFusion ~\cite{dreamfusion} introduced score distillation sampling (SDS) to train a NeRF ~\cite{nerf} using pretrained text-to-2D diffusion models ~\cite{imagen,stablediffusion}. Later methods proposed enhancements over SDS to improve the texture quality of the generated 3D models ~\cite{magic3D,texture3D,fantasia,mvdream,augmentedembed,dreamdissector,scaledreamer,hifa,sdi,cfd,dreamdpo}. To improve the appearance of the 3D model, ProlificDreamer ~\cite{prolificdreamer} proposed a variational score distillation approach, while IT3D ~\cite{it3D} proposed to combine GAN ~\cite{gan} loss with the SDS. There have also been methods for fast generation of text-to-3D models such as Latent Nerf ~\cite{latentnerf}, DreamGaussian ~\cite{dreamgaussian} and TPA3D ~\cite{tpa3D}. However, most of these methods fail to generate view consistent 3D models, resulting in ``Janus artifacts", which results in anatomically inconsistent human and hand models. 

\noindent\textbf{Text-to-human model generation: } There have been several methods to generate 3D human avatars from text prompts. Earlier methods proposed GAN based formulations to generate human avatars ~\cite{eva3D,lsvgan,getavatar,get3Dhuman}. Recent methods propose to combine score distillation sampling with parametric human models such as SMPL ~\cite{smpl}, SMPLX  ~\cite{smplx} and Neus ~\cite{neus} to generate 3D models consistent with human anatomy. Methods such as DreamAvatar ~\cite{dreamavatar}, AvatarCraft ~\cite{avatarcraft}, TADA ~\cite{tada} and AvatarCLIP ~\cite{avatarclip} propose to use such human models along with additional techniques such as CLIP loss and hierarchical rendering. Methods such as DreamWaltz ~\cite{dreamwaltz}, DreamHuman ~\cite{dreamhuman}, Avatarverse ~\cite{avatarverse} and Human Norm ~\cite{humannorm} propose to enhance the diffusion models using various additional controls such as human pose, depth map, normal map and dense mesh. Another method, AvatarBooth ~\cite{avatarbooth} proposes separate models to generate face and body. Recently HumanGaussian ~\cite{humangaussian} proposed a Gaussian Splatting ~\cite{gausssplat} approach for text-to-3D human avatar generation. However, we observed that these methods generate hand regions with very less details. To the best of our knowledge, there is no existing method for zero-shot 3D hand model generation from text prompts. 


\section{Background}

\noindent\textbf{Differentiable 3D representations} such as Neural Radiance Fields (NeRF) enable 3D model reconstruction from a set of multi-view 2D input images using a neural network ~\cite{nerf}. This can be formulated as: $g_\theta(\mathbf{k}) \mapsto (\sigma,c)$, where $g_\theta$ denotes the NeRF model, and $\sigma$ and $c$ denotes the volumetric density and color respectively, at the 3D location $\mathbf{k}$. Other representations such as Gaussian splatting proposes to model the 3D scene using learnable gaussians. Once the 3D representation is trained, an image can be rendered from any novel view point ($v$) using volumetric rendering along a batch of ray terminating at the required image plane or using rasterization techniques. 

\noindent\textbf{Score Distillation Sampling (SDS),} introduced by DreamFusion ~\cite{dreamfusion}, enables generation of 3D models from text prompts using knowledge distillation from pretrained 2D diffusion models into a 3D representation. Let $\mathbf{z}$ denote the latent space representation of an image $(\mathbf{x})$ obtained from the 3D model $g_\theta$ for view $v$. Then gradient of SDS loss is defined as:

\begin{equation}
\nabla_\theta\mathcal{L}_{SDS} = \mathbb{E}_{t,\epsilon}\left[w(t)(\epsilon_\phi(\mathbf{z}_t;y,t) - \epsilon)\frac{\partial \mathbf{z}}{\partial \theta}\right]
\label{eq:sds}
\end{equation}

Here, $\epsilon_\phi$ denotes a pretrained diffusion model parameterized by $\phi$; $\epsilon$ denotes the noise predicted by $\epsilon_\phi$; $t$ denotes the timestep of the forward diffusion process and $y$ denotes a conditional input, which is text prompt, in this case. $w(t)$ denotes a weighing parameter and $\mathbf{z}_t$ denotes the noisy signal after performing forward process for $t$ timesteps. The obtained gradient is then back-propagated through the differentiable 3D model to update the parameters $\theta$.

\begin{figure*}[tb]
  \centering
  \includegraphics[width=\textwidth]{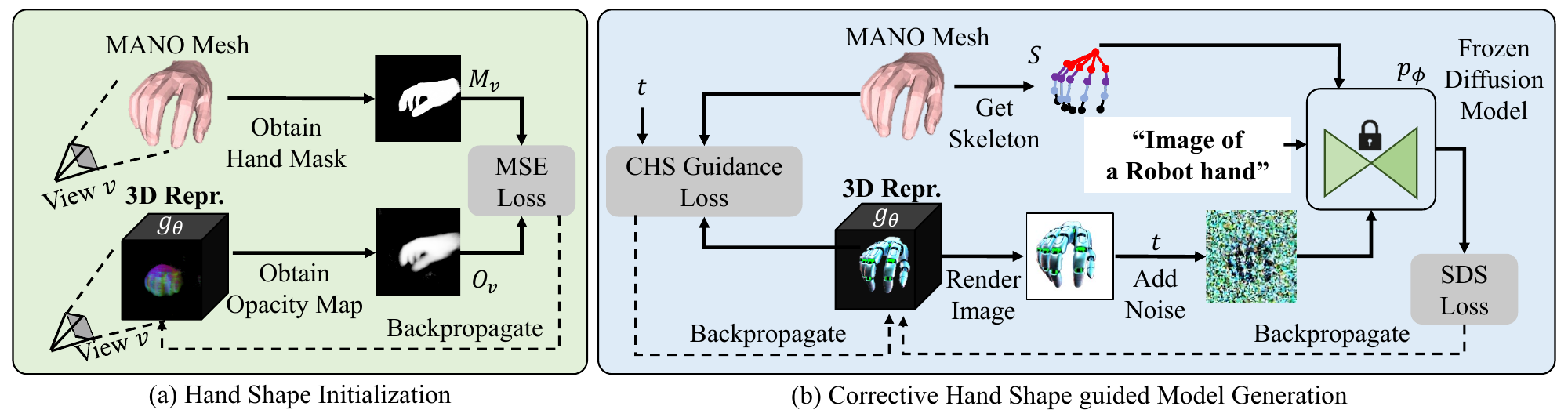}
  \caption{Overview of HandDreamer. Our method generates 3D hand models from text prompts in 2 stages: (a) Hand shape initialization using MANO mesh; (b) Hand model generation using skeleton and Corrective Hand Shape (CHS) guidance loss.}
  \label{fig:method}
\end{figure*}

\section{Investigating the origin of view-inconsistencies}
\label{sec:theory}
In this section, we investigate the origin of view-inconsistencies (``Janus" artifacts) in Score Distillation Sampling. Previous methods ~\cite{modeCollapse, dreamtime} suggested mode collapse as a reason for Janus artifacts. Particularly, ESD ~\cite{modeCollapse} postulated that SDS optimizes each view from $g_\theta$ to have maximum log likelihood on $p_\phi(\mathbf{z_0|y'})$, leading the views to collapse to the mode of $p_\phi(\mathbf{z_0|y'})$. Here, $y'=y+y_v$, where $y_v$ is a view dependent prompt $\in \{front, back, side, top, bottom\}$. 

However, our analysis suggests that avoiding mode collapse does not solve the view-inconsistent Janus artifacts, especially for highly articulatable objects such as hands. This is because the probability landscape of $p_\phi(z_t|y')$ consists of several probable modes due to the huge variation in camera poses and hand articulations (Fig. \ref{fig:modeTheory} (a)). The SDS optimization does not ensure that each view converges to the ``correct" mode. We provide a toy example in Fig. \ref{fig:modeTheory} (b) to explain this. The blue region denotes the distribution $p_\phi(z_t|y')$ where $t$ is close to zero and $y'=y + ``front"$. Let $\mu_1$ and $\mu_2$ denote two probable modes of this distribution for the text prompt $y'$. Let $z_t$ denote an initial condition for view $v$ obtained using a randomized initialization of the 3D representation. Let $s_\phi(z_t) \sim \nabla_{z_t}\log p_\phi(z_t)$ denote the score of the probability function $p_\phi$. ~\cite{dreamfusion} shows that SDS optimization follows the path that minimizes the estimated absolute score value ($s_\phi(z_t)$). However, at a low value of timestep $t_{low}$, $z_t^v$ is out of distribution with very low probability value. Hence $s_\phi$ is not reliable in this region, necessitating evaluation at a higher timestep $t_{high}$ during initial iterations, by making the distribution more noisy (orange region). However, the noise addition is stochastic and hence $z_t^v$ gets pushed towards different modes for very similar views, resulting in view-inconsistencies and Janus artifacts. These are further aggravated when the number of probable modes are high for a given $y'$. This is especially evident for highly articulatable objects such as hands, where the number of modes are tremendously high due to large variations in hand poses. Conversely, our analysis also suggests that view-inconsistencies can be reduced using an initialization with a low expected absolute $s_\phi$ with respect to each of the ``correct" (ground-truth) modes for all $v \in V$ for low values of $t$. To gain more insights on designing SDS without view-inconsistencies for hands we present the following theorem. 

\noindent \textbf{Theorem 1.} \textit{Let $x_{latent}^v$ and $x_{init}^v$ denote the set of views rendered from an ideal latent 3D model ($m_{3D}^{latent}$) and an initial 3D model ($m_{3D}^{init}$) respectively. Then the expected absolute score of $m_{3D}^{init}$ w.r.t $m_{3D}^{latent}$, denoted as $|S_\phi|$ is:}
\begin{equation}
\left|\mathbb{E}_{v}\left[\frac{-\sqrt{\bar{\alpha_t}}}{1-\bar{\alpha_t}}\left[(\mathcal{E}(x_{init}^v)-\mathcal{E}(x_{latent}^v)) + \frac{\sqrt{1 - \bar{\alpha_t}}}{\sqrt{\bar{\alpha_t}}}\epsilon\right]\right]\right|
\label{eq:theorem}
\end{equation}

\noindent Here $\mathcal{E}$(.) is the encoder of Stable Diffusion and $\bar\alpha_t$ denotes forward noise parameters of the diffusion model, with $\bar{\alpha_t}\rightarrow 0$ as $t$ increases. We provide the proof of the theorem in Appendix A.1. Since $\epsilon\sim\mathcal{N}(0,\mathbf{I})$, the second term of Eq. \ref{eq:theorem} approaches zero for a large number of samples. Hence a low score initial model can be obtained by either increasing $t$ or by choosing $m_{3D}^{init}$ such that $|z_{init}^v - z_{latent}^v|$ is low for all $v \in \{V\}$, where $z=\mathcal{E}(x)$. However, increasing $t$ will reduce the score w.r.t to other modes present in $p_\phi$ as well, since $|S_\phi|$ becomes independent of the ideal latent mode ($z_{latent}$) as $\bar{\alpha_t} \rightarrow 0$. Since this would result in view-inconsistencies, an ideal initial condition should minimize $|z_{init}^v - z_{latent}^v|$. i.e, be semantically closer to the required view-consistent latent 3D ground-truth. 

We carefully design our methodology for text-to-3D hand generation based on these analyses. Firstly, we propose to initialize our 3D representation using a MANO hand model so that $m_{3D}^{init}$ is semantically and geometrically close to the ideal view-consistent hand model, compared to an standard spherical initialization. Secondly, we propose to minimize the number of probable modes for each view by providing an additional condition which encodes both the viewpoint and hand pose information to the diffusion model. We empirically observe that the 2D projection of 3D hand skeleton onto view $v$ effectively embeds both these information. The impact of these steps on the gradient ($\frac{\partial L_{sds}}{\partial z_t}$) is shown in Fig. \ref{fig:modeTheory} (c,d). At lower $t$ the scores of random initialization is unreliable resulting in less informative gradients (Fig. \ref{fig:modeTheory}(c - left)). At larger $t$, random initialization leads to gradients that pushes the optimization towards diverse modes for the same view points for different $\epsilon \sim \mathcal{N}(0,\mathbf{I})$, resulting in view-inconsistencies (Fig. \ref{fig:modeTheory}(d - left)). On the other hand, MANO based initialization results in consistent gradients for the same viewpoint at both $t_{low}$ and $t_{high}$. 

While these steps helped address view-inconsistencies by improving mode convergence, we  observed geometric distortions in the later stages of optimization. This is especially pronounced in the side views of hands where there is significant self-occlusion of fingers (Fig. \ref{fig:abl}-c). To address this issue, we further propose a constant mode correction using a novel corrective hand shape guidance loss. Our algorithm is detailed in the next section. 

\begin{figure*}[tb]
  \centering
  \includegraphics[width=\textwidth]{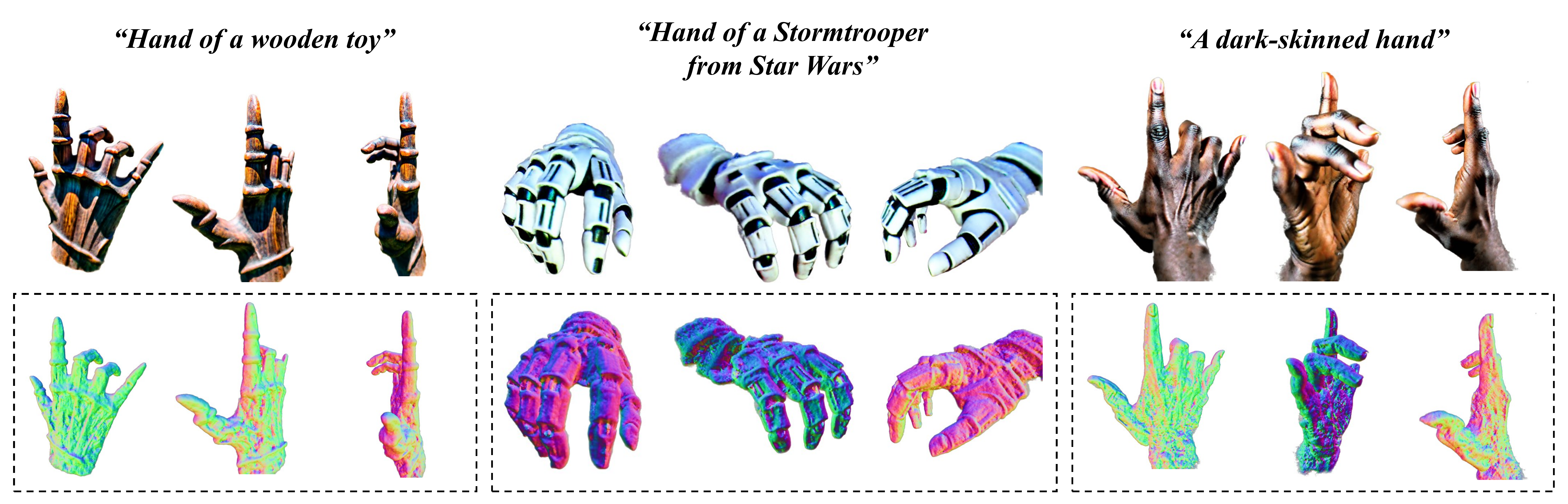}
  \caption{Our method generates 3D hand models with detailed texture and view-consistent geometry. Surface maps provided inset}
  \label{fig:dreamOutputs}
\end{figure*}

\section{HandDreamer}
Fig. \ref{fig:method} provides an overview of our proposed HandDreamer method. We utilize a NeRF based 3D representation in this pipeline. We first  initialize the volumetric density of the NeRF using a MANO hand model. For this, we minimize the error between opacity maps of the NeRF and that of a MANO hand mesh in the required hand pose. Next, we perform SDS guided by the text prompt, 2D hand skeleton rendered from multiple views and a corrective hand shape guidance loss. Detailed description of our algorithm is provided in the following subsections.

\subsection{Hand Shape Initialization}

We initialize the volumetric density of the NeRF using a MANO hand model in the desired hand pose, thereby providing a robust geometric prior for the latter stages. During the training process, we minimize the error between hand silhouette masks obtained from various views of the MANO model, and opacity masks obtained from the NeRF. For a given view $v$, we obtain the opacity mask of the NeRF as follows.

Let \(\mathbf{r}(t) = \mathbf{p} + t\mathbf{d}\) denote a ray originating from pixel $\mathbf{p}$ in the direction $\mathbf{d}$. We sample $T$ points along the ray. The opacity value for $\mathbf{p}$ is given as ~\cite{nerf}:
\begin{equation}
O_{v,\mathbf{p}} = \sum_{i=1}^{T}\left(\prod_{j=1}^{i-1} \exp\left(-\sigma_j \delta_j\right)\right) \left(1 - \exp\left(-\sigma_i \delta_i\right)\right)
\label{eq:weight}
\end{equation}
where \(\sigma_i\) denotes the volumetric density at the \(i\)-th sample and \(\delta_i\) is the distance between consecutive samples. We perform min-max normalization over the opacity map, assigning opaque pixels to 1.0 and transparent pixels to 0.0. To initialize the NeRFs, we minimize the L2 distance between the opacity maps ($O_v$) and the ground truth silhouette mask from MANO $(M_v)$. 

\subsection{Corrective Hand Shape guided Model Generation}
\noindent \textbf{Score Distillation Sampling:} Once the 3D representation is initialized, we perform a hand skeleton-guided score distillation sampling (SDS) to generate the final 3D hand model pertaining to the provided text prompt. The 2D projection of the 3D hand skeleton embeds both the view point and the hand pose in a single image, which helps minimize the modes of the probability landscape $p_\phi(z|y')$. We implement this using a ControlNet ~\cite{controlnet} trained using hand skeleton as control. We also use square-root timestep annealing proposed by ~\cite{hifa} so that the sampled noise gradually decreases as the optimization proceeds. In each training iteration, we render an image $I$ from the NeRF. Next we add noise corresponding to the timestep annealing schedule ($t$), resulting in a noisy image $(I_t)$. Simultaneously, we extract hand keypoints from the MANO mesh for the same viewpoint to obtain the hand skeleton $(S)$. The noisy image \(I_t\), the hand skeleton \(S\), and the text prompt \(y\) are then provided to the frozen ControlNet to estimate the noise present in \(I_t\) at noise-timestep \(t\). We compute the gradient using score distillation sampling (SDS) to update the weights of the NeRF as follows:

\begin{equation}
\nabla_\theta\mathcal{L}_{SDS} = \mathbb{E}_{t,\epsilon}\left[w(t)(\epsilon_\phi(\mathbf{I}_t;y,t,S) - \epsilon)\frac{\partial \mathbf{I}}{\partial \theta}\right]
\label{eq:sds2}
\end{equation}

\noindent \textbf{Corrective Hand Shape Guidance:} We observed that training the 3D model using only SDS loss often results in geometric distortions such as inconsistent thickness especially in the side views of the hand. We postulate that this arises due to significant self-occlusion of fingers in these views. To address this challenge, we propose a novel Corrective Hand Shape (CHS) loss using additional guidance from MANO hand shape prior. During every iteration of SDS, we also ``correct" the hand shape by minimizing the L2 distance between the opacity of the NeRF and silhouette mask from MANO. This ensures that views from the 3D model do not deviate too much from the required geometric modes. We also observed that SDS tends to perform more geometric updates at higher noise $t$ and more texture updates at lower $t$ (Supplementary A.3). Hence we also anneal the CHS loss to provide more weight at higher $t$. The final CHS loss is given as follows:

\begin{equation}
\mathcal{L}_{chs}(t) = \lambda_t^{chs}.\frac{1}{|\mathcal{V}|}\sum_{v\in \mathcal{V}}\|O_v - M_v\|_2
\label{eq:chsLoss}
\end{equation}

\noindent where $\lambda_t^{chs}$ is annealed as follows:

\begin{equation}
\lambda_t = \lambda_{max}^{chs}\left[\frac{t-t_{min}}{t_{max}-t_{min}}\right] + \lambda_{min}^{chs}\left[\frac{t_{max}-t}{t_{max}-t_{min}}\right]
\label{eq:lambdaChs}
\end{equation}

\noindent Eq. \ref{eq:lambdaChs} is derived using the definition of square-root timestep annealing ~\cite{hifa} (Please refer to supplementary A.1). We have empirically chosen $\lambda_{max}$, $\lambda_{min}$, $t_{max}$ and $t_{min}$ as 15000, 1000, 600 and 300 respectively. 

\noindent \textbf{Additional Loss Functions:} We also use image loss ($\mathcal{L}_{img}$) and z-variance ($\mathcal{L}_{zvar}$) losses proposed by ~\cite{hifa} to further stabilize the training and generate sharp and high-fidelity outputs. While $\mathcal{L}_{img}$ ensures that the 3D model has high fidelity in both image and latent spaces, $\mathcal{L}_{zvar}$ densifies the volume density to ensure the the surface of the 3D model is sharp and well-defined. The final loss function is given as:

\begin{equation}
\mathcal{L} = \lambda_{sds}.\mathcal{L}_{sds} + \lambda_t^{chs}.\mathcal{L}_{chs}(t) + \lambda_{img}.\mathcal{L}_{img}  + \lambda_{zvar}.\mathcal{L}_{zvar}
\label{eq:lossFull}
\end{equation}

\noindent where $\lambda_{sds}$, $\lambda_{img}$ and $\lambda_{zvar}$ are empirically chosen as 1, 0.01 and 100 respectively.

\begin{figure*}[tb]
  \centering
  \includegraphics[width=\textwidth]{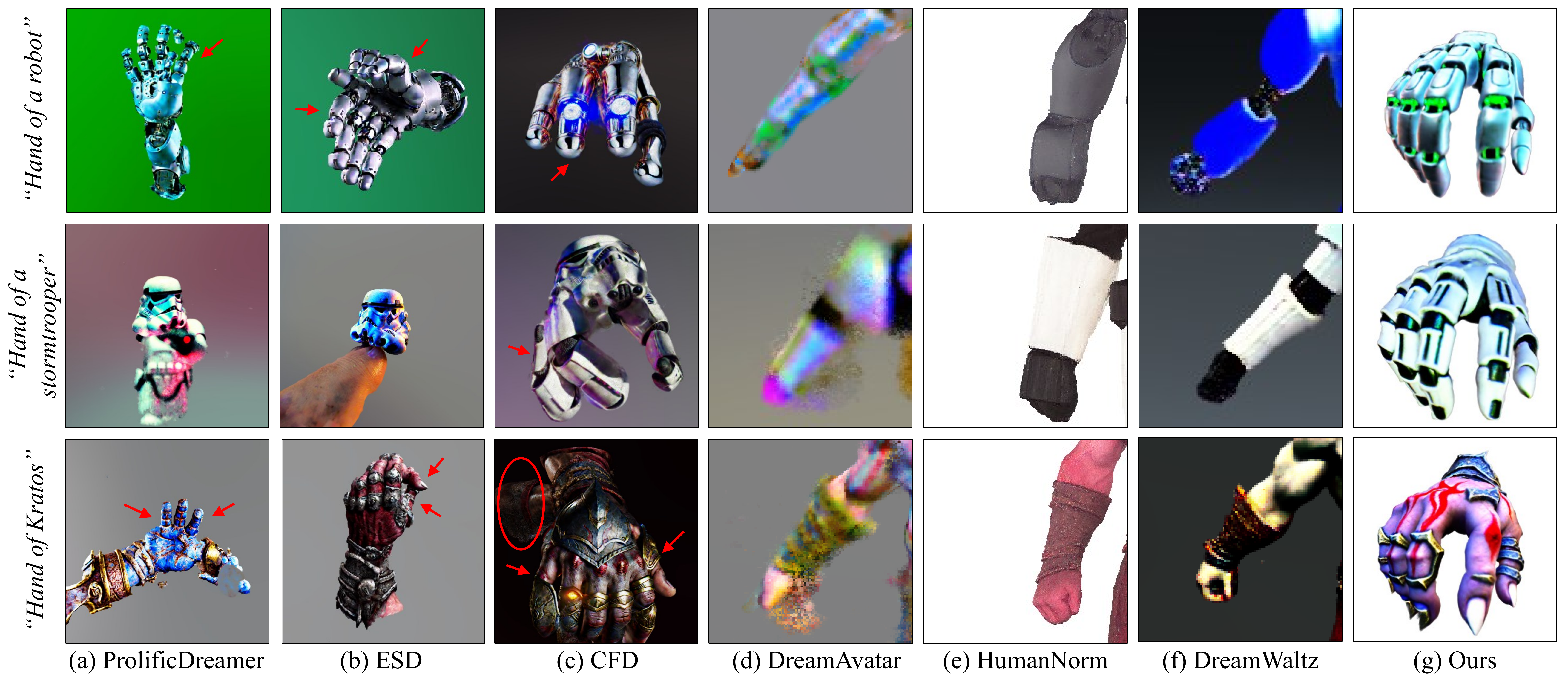}
  \caption{Comparison against state-of-the-art text-to-3D methods. Janus artifacts and inconsistent fingers shown in red arrows and circle (a-c). Text-to-human methods (d-f) generates hands with very less details. Our method generates better 3D hand models with consistent geometry and details.}
  \label{fig:results}
\end{figure*}

\begin{figure}[tb]
  \centering
  \includegraphics[width=\columnwidth]{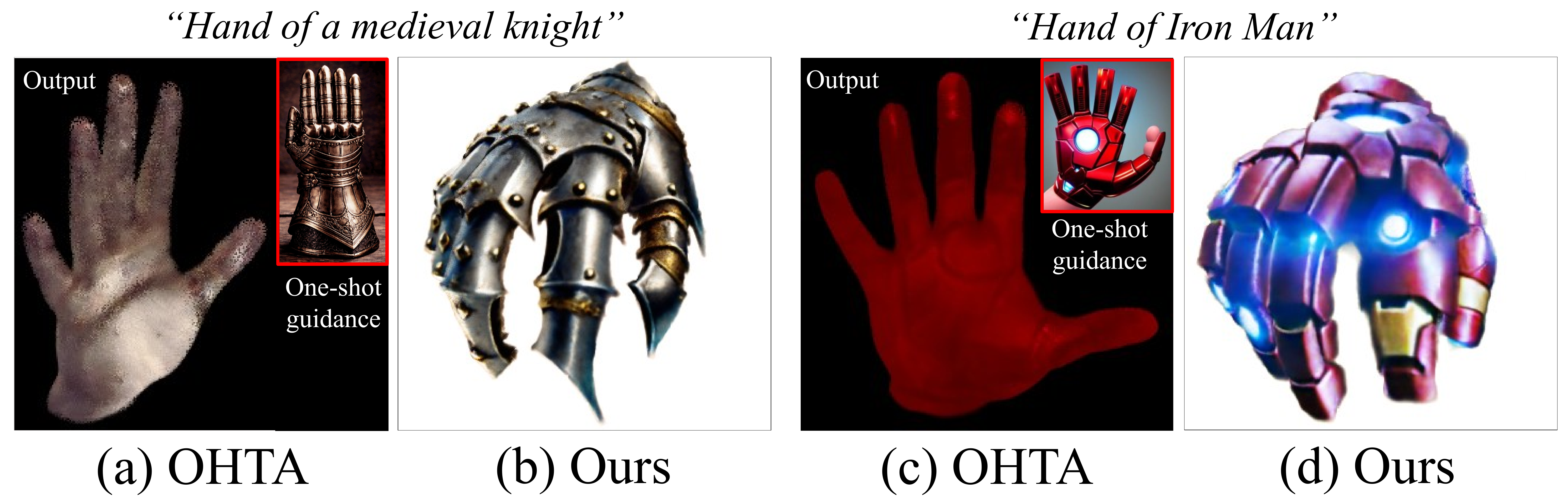}
  \caption{Comparison against one-shot method OHTA. Our method generates better textures and diverse geometry.}
  \label{fig:ohtaComp}
\end{figure}

\section{Experiments and Results}

\subsection{Implementation Details}

We use a NeRF based 3D representation ~\cite{instantngp} and Stable Diffusion 1.5 ~\cite{stablediffusion} with Controlnet 1.1 ~\cite{controlnet} as our 2D generative model. Our method is implemented using PyTorch and Threestudio framework ~\cite{threestudio}. We perform our experiments on a NVIDIA RTX A6000 GPU with 48 GB of RAM. The hand shape initialization and the CHS guided SDS stages converges in 2000 and 8000 iterations respectively. The hand shape initialization stage, which takes $\sim$ 15 minutes, needs to be executed only once and can be reused for any prompt. The CHS guided SDS stages takes $\sim$ 45 minutes for convergence and consumes $\sim$ 30 GB of GPU RAM. Additional details are provided in the supplementary material.

\subsection{Qualitative Evaluation}

We provide examples of 3D hand models generated by the proposed HandDreamer method in Fig. \ref{fig:dreamOutputs}. We show rendered images from three different viewpoints with diverse articulations, along with surface maps. It can be seen that our method generates high-fidelity 3D hand models with diverse and view-consistent geometry with detailed texture. We provide more multi-view examples in the supplementary material. 

\noindent\textbf{Comparisons against state-of-the-art: } We provide qualitative comparisons against state-of-the art methods in text-to-3D: ProlificDreamer ~\cite{prolificdreamer}, ESD ~\cite{modeCollapse} and CFD ~\cite{cfd}; and text-to-3D human: DreamWaltz ~\cite{dreamwaltz}, DreamAvatar ~\cite{dreamavatar}, HumanNorm ~\cite{humannorm} in Fig. \ref{fig:results}. The results of these methods are obtained using the codes provided by the authors.  It can be seen that ProlificDreamer and ESD suffer from Janus artifact resulting in fingers protruding at unnatural locations on the hand (Fig. \ref{fig:results}(a,b)); and CFD results in anatomically incorrect hand models with incorrect number of fingers (Fig. \ref{fig:results}(c)). On the other hand, text-to-human methods (Fig. \ref{fig:results}(d-f)) generate images with very little details in the hand regions. In contrast, our method (Fig. \ref{fig:results}(g)) is able to generate superior hand models with detailed texture and accurate geometry for a variety of prompts. We also provide comparisons against OHTA, a one-shot image generation method, in Fig. \ref{fig:ohtaComp}. We generated an input image for this method using Stable Diffusion 1.5. It can be seen that this method, which obtains texture using a learnt database, doesn't adapt very well to generic hands, and results in lack of geometric diversity and textural fidelity (Fig. \ref{fig:ohtaComp} (a,c)). On the other hand, our method generates superior results with intricate details and diverse textures ((Fig. \ref{fig:ohtaComp} (b,d)).

\begin{figure}[tb]
  \centering
  \includegraphics[width=\columnwidth]{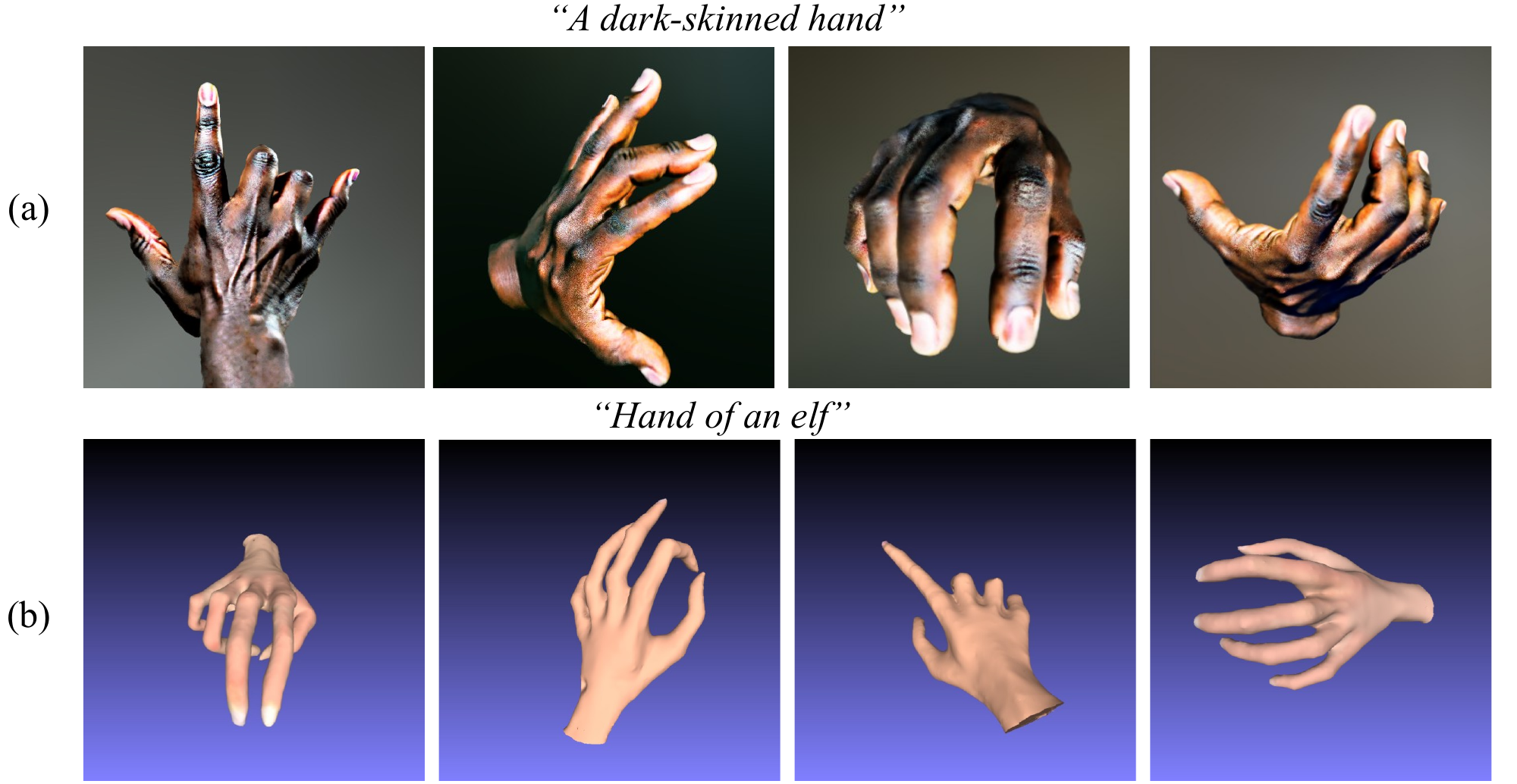}
  \caption{Our method supports hand articulations for diverse hand poses.(a) Obtained using different MANO parameters. (b) Obtained using rigging of exported mesh.}
  \label{fig:art}
  \vspace{-0.5cm}
\end{figure}

\noindent\textbf{Articulation:} Our method can generate hand poses with diverse articulations (Fig. \ref{fig:art}), either by changing the MANO parameters, or by using downstream graphics pipelines. In Fig. \ref{fig:art} (a), we show an example wherein diverse poses are obtained by changing the MANO parameters used in the algorithm. Further, the 3D hand models generated by our method can be exported to meshes for downstream graphics tasks such as rendering and animation. We use ~\cite{threestudio} to export the 3D hands into a high quality mesh with $\sim$ 300k vertices. Next, we rig a hand skeleton onto the mesh for articulation. The rigged model can be animated or articulated using any driving pose. We have provided examples of articulated rigged meshes for complex poses in Fig. \ref{fig:art}(b).

\setlength{\tabcolsep}{2pt}
\begin{table}[t]
\begin{center}

\begin{tabular}{cccc}
\hline
	Method & CLIP L14 $\uparrow$ & FID $\downarrow$ & HPSv2 $\uparrow$\\
\hline  \\[-1.5ex]
DreamFusion'22 ~\cite{dreamfusion} &  25.12 & 344.19 & 0.187 \\
LatentNerf'23 ~\cite{latentnerf} & 24.34 & 316.42 & 0.189\\
Fantasia3D'23 ~\cite{fantasia}  & 20.93 & 329.31 & 0.198\\
DreamWaltz'23 ~\cite{dreamwaltz} & 23.96 & 265.11 & 0.222\\
DreamAvatar'24 ~\cite{dreamavatar} & 20.02 & 329.85 & 0.215\\
HumanNorm'24 ~\cite{humannorm} & 23.01 & 327.42 & 0.177\\
SDI'24 ~\cite{sdi}& 26.32 & 297.12 & 0.192 \\
OHTA'24 ~\cite{ohta}& 22.59 & 467.51 & 0.181 \\
CFD'25 ~\cite{cfd}& \underline{26.62} & \underline{262.83} & \underline{0.223} \\

\textbf{HandDreamer (Ours)}  & \textbf{28.63} & \textbf{254.62} & \textbf{0.241} \\
\hline
\end{tabular}
\caption{Quantitative comparisons. Our method outperforms the other methods on all the metrics.}
\vspace{-0.75cm}
\end{center}
\label{tab:sota1}
\end{table}

\subsection{Quantitative Evaluation}

\noindent\textbf{Objective Evaluation: } We also provide quantitative comparisons using CLIP L14 score ~\cite{clipscore}, Fréchet inception distance (FID) and Human Preference Score (HPSv2) ~\cite{hpsv2} in Tab. 1. We create a test set comprising of 5400 images rendered from equidistant viewpoints from 45 different 3D models for this evaluation. While CLIP score measures the similarity between the text prompt and the rendered images, FID quantifies their realism. On the other hand HPSv2 is a recently introduced metric that quantifies the human preference of each image. Similar to ~\cite{humannorm}, we obtain ground-truth images for FID using a pre-trained diffusion model ~\cite{controlnet}. We evaluated our method against a comprehensive set of text-to-3D methods: DreamFusion~\cite{dreamfusion}, LatentNerf~\cite{latentnerf}, Fantasia3D~\cite{fantasia}, SDI~\cite{sdi}, CFD~\cite{cfd}, DreamWaltz~\cite{dreamwaltz}, DreamAvatar~\cite{dreamavatar} and HumanNorm~\cite{humannorm} and OHTA ~\cite{ohta}.  It can be seen that our method significantly outperforms the existing methods on all the metrics, indicating a higher semantic similarity, realism and human preference. 

\begin{figure}[tb]
  \centering
  \includegraphics[width=\columnwidth]{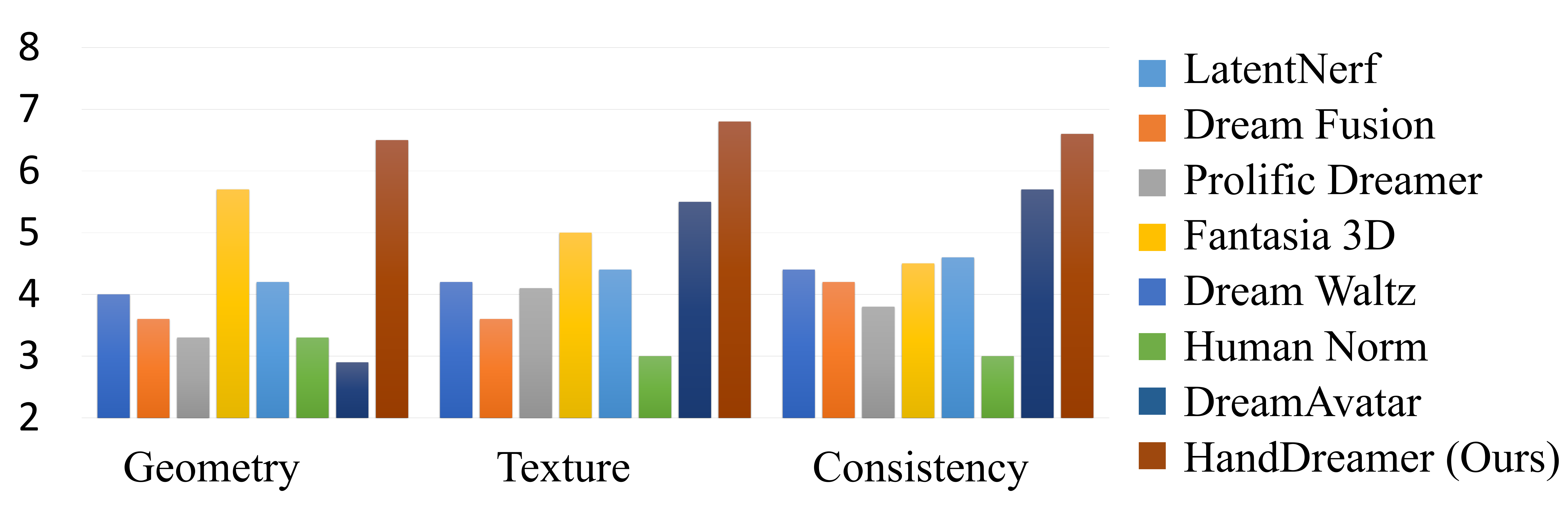}
  \caption{User preference study. Higher value is better. Our method scores highest compared to the other methods.}
  \label{fig:userstudy}
\end{figure}

\noindent\textbf{User Studies: } We also conducted studies to evaluate user preference compared to the existing methods. We evaluated the geometry, texture and consistency of the generated 3D hand model with the text prompts. We asked 50 volunteers, aged 21 to 35, to rate the 3D models generated by 8 different methods on 30 different prompts on a scale of 1 (worst) to 8 (best), in a blind and randomized manner. For each of the methods, the user was given a video rendered from the 3D models showing multiple viewpoints. The results summarized in Fig. \ref{fig:userstudy} shows that the proposed HandDreamer method is preferred on an average on all the three criteria. 

\begin{figure}[tb]
  \centering
  \includegraphics[width=\columnwidth]{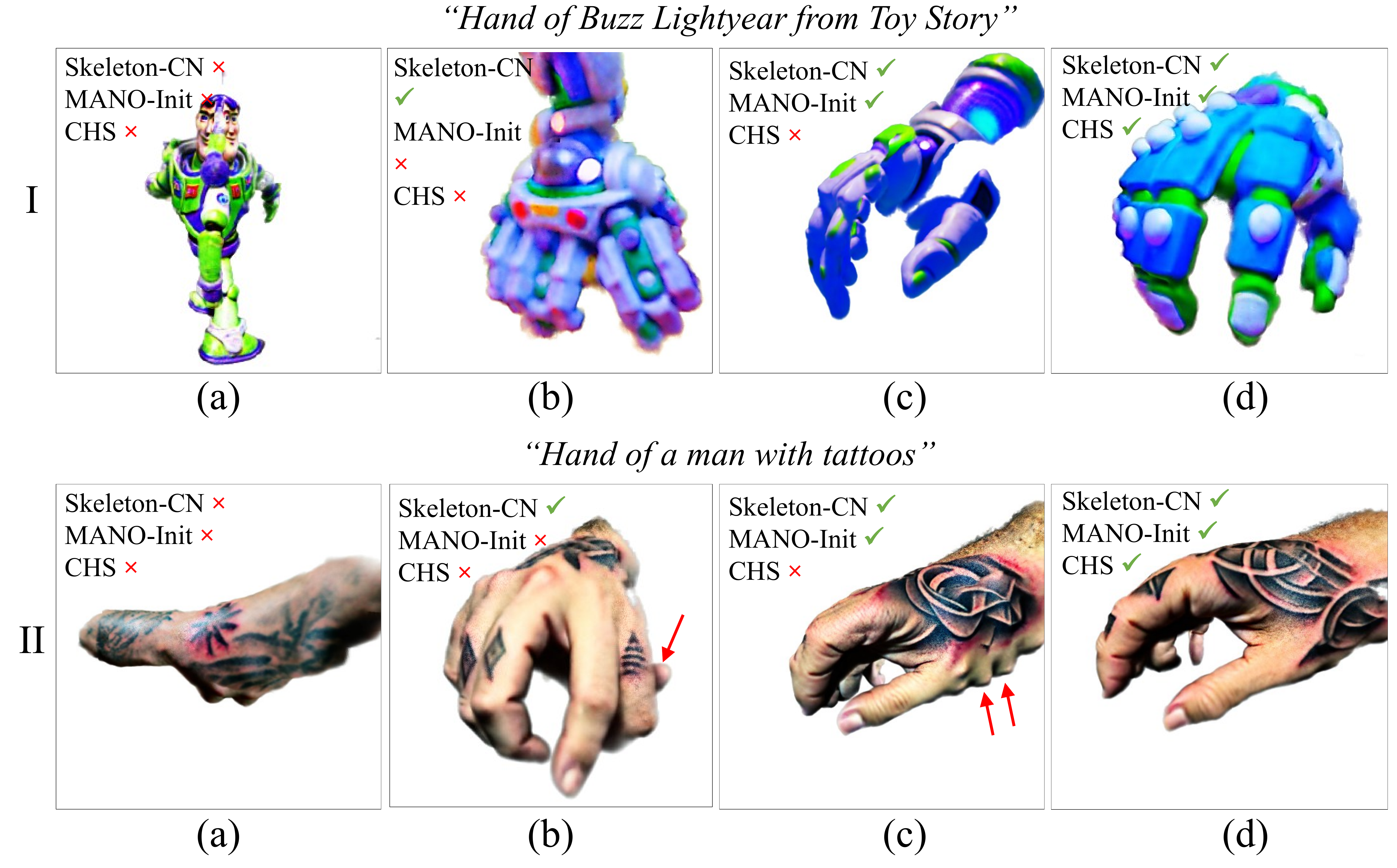}
  \caption{Ablation studies. (a) Removing all components fails to generate hand structure. (b) Using only CN causes incorrect geometry. (c) Removing CHS causes distortions in side views. (d) Using all components generates best results.}
  \label{fig:abl}
\end{figure}

\setlength{\tabcolsep}{6pt}
\begin{table}[t]
\begin{center}
\begin{tabular}{cccc}
\hline
	Skeleton-CN & MANO Init & CHS loss & CLIP L14$\uparrow$ \\
\hline  \\[-1.5ex]
$\times$ & $\times$ & $\times$  & 26.40  \\
$\surd$ & $\times$ & $\times$   & 26.67  \\
$\surd$ & $\surd$ & $\times$  & 28.48 \\
$\times$ & $\surd$ & $\surd$ & 27.07 \\
$\surd$ & $\times$ & $\surd$ & 28.02 \\
$\surd$ & $\surd$ & $\surd$   & \textbf{28.63}  \\
\hline
\end{tabular}
\caption{Ablation Studies. Our method using all the components generates the best results.}
\vspace{-0.75cm}
\end{center}
\label{tab:sota1}
\end{table}

\subsection{Ablation Studies}

We also analyse the impact of the core components of our algorithm: skeleton based ControlNet, MANO initialization and Corrective Hand Shape (CHS) guidance loss function. The results of these studies are provided in Fig. \ref{fig:abl} and Tab. 2. We use the same set of 45 prompts used for comparative studies here. For the first experiment, we removed  all the three components. From Fig. \ref{fig:abl} (a), it can be seen that this results in severe Janus artifacts (row I), geometric distortions (row II) and ignores the requirement to generate the hands (row I). From Tab. 2 (row 1), it can be seen that the the CLIP score is significantly degraded when all the 3 components are removed. 

Next, we introduced skeleton based ControlNet (Skeleton-CN) (Fig. \ref{fig:abl} (b)). While this helped to generate a 3D model with the shape of a hand, the geometric details are inaccurate. For example, it can be seen that this results in erroneous number of fingers (Fig. \ref{fig:abl} (b), Row 2, red arrows). The CLIP score provided in Tab. 2 (row 2) further validate this observation. 

Next, we introduced MANO based initialization ((Fig. \ref{fig:abl} (c)). It can be seen that this resulted in a 3D model with higher fidelity. However, this also results in geometric degradations, especially in the side poses of the hand, leading to disconnected segments (Fig. \ref{fig:abl} (c), row 1), and unnatural artifacts (Fig. \ref{fig:abl} (c), row 2, red arrow). From Tab. 2 (row 3), it can be seen that using both MANO initialization and skeleton based controlnet significantly improves the CLIP score. However, this is still lower than the full method (Tab. 2, row 6). Further, from Fig. \ref{fig:abl} (d), it can be seen that the full method with all the components results in the best outputs with consistent views, high fidelity, and geometric accuracy. 

We also quantitatively analysed the impact of removing only the skeleton based controlnet (Tab. 2, row 4) and the MANO initialization (Tab. 2, row 5). It can be seen that both these experiments resulted in a lower CLIP score compared to the full method (Tab. 2, row 6), signifying the importance of geometric priors on output quality.

\section{Conclusion}

In this paper, we present HandDreamer, the first method for zero-shot 3D hand model generation from text prompts. We observe that view-inconsistencies in SDS is primarily caused due to ambiguity of modes in the probability distribution defined by the text prompt. To alleviate this problem, we propose to reduce the number of plausible modes using a hand skeleton based condition. Further, we also propose to use a low-score initialization using a MANO hand model. Additionally, we propose a novel corrective hand shape loss to ensure that the hand geometry does not get degraded during SDS. Through extensive experimentation, we show that our method achieves state-of-the-art results in 3D hand generation, both quantitatively and qualitatively.  We also show that our method can generate a wide variety of hand geometry with high quality texture, without issues such as Janus artifacts. Our method enables creation of high quality hand avatars with ease, which is highly useful in applications such as virtual reality and gaming. 

\noindent\textbf{Limitations:} Similar to other SDS based methods, our method can potentially inherit biases of the pre-trained diffusion model. Further, articulation of our optimized 3D hand models requires exporting to mesh and rigging. In the future, we will explore automated ways for articulation.

{
    \small
    \bibliographystyle{ieeenat_fullname}
    \bibliography{main}
}
\newpage
\section{Supplementary}

We provide proofs and derivations for the results presented in the main paper in section A.1 We provide additional implementation details in section A.2. Additional studies are provided in section A.3, additional quantitative results in section A.4 and more results in section A.5. We also provide multi-view videos in the multimedia supplementary attachment. 

\section{A.1 Proofs and Derivations}

In this section, we provide proofs and derivations for the theorems and definitions defined in the main paper. 

\subsection{Proof for Theorem 1}

\noindent \textbf{Theorem 1.} \textit{Let $x_{latent}^v$ and $x_{init}^v$ denote the set of views rendered from an ideal latent 3D model ($m_{3D}^{latent}$) and an initial 3D model ($m_{3D}^{init}$) respectively. Then the expected absolute score of $m_{3D}^{init}$ w.r.t $m_{3D}^{latent}$ is:}
\begin{equation}
|S_\phi| = \left|\mathbb{E}_{v}\left[\frac{-\sqrt{\bar{\alpha_t}}}{1-\bar{\alpha_t}}\left[(\mathcal{E}(x_{init}^v)-\mathcal{E}(x_{latent}^v)) + \frac{\sqrt{1 - \bar{\alpha_t}}}{\sqrt{\bar{\alpha_t}}}\epsilon\right]\right]\right|
\label{eq:theorem}
\end{equation}

\noindent where $\mathcal{E}$(.) is the encoder of Stable Diffusion and $\bar\alpha_t$ denotes forward noise parameters of the diffusion model. Let $z_t^{latent} = \sqrt{\bar{\alpha_t}}\mathcal{E}(x_{latent}^v)$ denote the mode towards which a view ($v$) should converge into. Since $p_\phi(z_t|y,t)$ is Gaussian, $z_t^{latent}$ is also the mode of a locally Gaussian distribution within $p_\phi$. Next we present the following lemma.

\noindent \textbf{Lemma 1.}  \textit{The score function of a multivariate Gaussian distribution $\mathcal{N}(\mathbf{x}; \mathbf{\mu}, \mathbf{\Sigma})$ is given as:}

\begin{equation}
s(\mathbf{x}) = -\Sigma^{-1}(\mathbf{x} - \mathbf{\mu})
\end{equation}

\noindent\textit{where, $\mathbf{\mu}$ and $\mathbf{\Sigma}$ denote the mean and covariance metrics respectively.}

\noindent\textit{Proof:}

\noindent Score of a probability distribution is defined as the derivative of the log likelihood as follows:

\begin{equation}
s(x) = \nabla_x\log p(x)
\end{equation}

\noindent Since $p(x)$ is a multivariate Gaussian with dimensionality $d$,

\small
\begin{equation}
\begin{split}
s(\mathbf{x}) & = \nabla_\mathbf{x}\log(\mathcal{N}(\mathbf{x}; \mathbf{\mu}, \mathbf{\Sigma})) \\
                     & = \nabla_\mathbf{x}\log\left[\frac{1}{(2\pi)^{d/2}|\Sigma|^{1/2}} \exp\left(-\frac{1}{2}(\mathbf{x} - \mathbf{\mu})^T\Sigma^{-1}(\mathbf{x} - \mathbf{\mu})\right)\right] \\
                     & = -\nabla_\mathbf{x}\left(\frac{d}{2}\log(2\pi) + \frac{1}{2}\log|\Sigma|\right) \\ & \hspace{1em}-\nabla_\mathbf{x}\left(-\frac{1}{2}(\mathbf{x} - \mathbf{\mu})^T\Sigma^{-1}(\mathbf{x} - \mathbf{\mu})\right)
\end{split}
\end{equation}
\normalsize

\noindent The first term of the above equation vanishes to 0 since it is independent of $\mathbf{x}$. The argument of the gradient of the second term is of quadratic form and hence it reduces to the following:

\begin{equation}
\begin{split}
s(\mathbf{x}) & = -\frac{1}{2}.2.\Sigma^{-1}(\mathbf{x} - \mathbf{\mu}) \\
                      & =  -\Sigma^{-1}(\mathbf{x} - \mathbf{\mu})         
\end{split}
\end{equation}

\noindent Next, we proceed to derive theorem 1 defined in Eq. \ref{eq:theorem}. 

\noindent\textit{Proof:}

\noindent Since $z_t^{latent}$ is the mode of locally isotropic Gaussian with $\Sigma=(\sqrt{1-\bar{\alpha_t}})\mathbf{I}$ at noise timestep $t$, the expected absolute score at any point $z_t$ w.r.t to the local probability distribution is given by lemma 1.

\begin{equation}
|\mathbb{E}_v[s(z_t)]| = \left|\mathbb{E}_v\left[-\Sigma^{-1} (z_t - \mathbf{\mu})\right]\right|
\end{equation}

\noindent Using $\Sigma=(\sqrt{1-\bar{\alpha}})\mathbf{I}$ and $\mathbf{\mu}=z_t^{latent}$, since the mean and mode are same for Gaussian,
\begin{equation}
|\mathbb{E}_v[s(z_t)]| = \left|\mathbb{E}_v\left[-\frac{1}{\sqrt{1-\bar{\alpha_t}}}(z_t -z_t^{latent})\right]\right|
\end{equation}

\noindent Using $z_t = \sqrt{\bar{\alpha_t}} z_0 + \sqrt{1-\bar{\alpha_t}}\epsilon$ and $z_t^{latent}=\sqrt{\bar{\alpha_t}}z_0^{latent}$,

\small
\begin{equation}
\begin{split}
|\mathbb{E}_v[s(z_t)]| & = \left|\mathbb{E}_v\left[-\frac{1}{\sqrt{1-\bar{\alpha_t}}}(z_t -z_t^{latent})\right]\right| \\
                                 & = \left|\mathbb{E}_v\left[-\frac{1}{\sqrt{1-\bar{\alpha_t}}}(\sqrt{\bar{\alpha_t}} z_0 + \sqrt{1-\bar{\alpha_t}}\epsilon - \sqrt{\bar{\alpha_t}} z_0^{latent})\right]\right| \\
                                 & = \left|\mathbb{E}_v\left[-\frac{1}{\sqrt{1-\bar{\alpha_t}}}(\sqrt{\bar{\alpha_t}} (z_0 -  z_0^{latent}) + \sqrt{1-\bar{\alpha_t}}\epsilon)\right]\right| \\
                                 & = \left|\mathbb{E}_v\left[\frac{-\sqrt{\bar{\alpha_t}} }{\sqrt{1-\bar{\alpha_t}}}(z_0 -  z_0^{latent}) + \frac{\sqrt{1-\bar{\alpha_t}}}{\sqrt{\bar{\alpha_t}}}\epsilon)\right]\right|
\end{split}
\end{equation}
\normalsize

\noindent Using a slight abuse of notations to denote $\mathbb{E}_v[s(z_t)]$ as $S_\phi$, the expected score of initial 3D model ($m_{3D}^{init}$), and reparameterizing $z_0=\mathcal{E}(x_{init}^v)$ and $z_0^{latent}=\mathcal{E}(x_{latent}^v)$ for view $v$, we arrive at theorem 1 as follows:

\begin{equation}
|S_\phi| = \left|\mathbb{E}_{v}\left[\frac{-\sqrt{\bar{\alpha_t}}}{1-\bar{\alpha_t}}\left[(\mathcal{E}(x_{init}^v)-\mathcal{E}(x_{latent}^v)) + \frac{\sqrt{1 - \bar{\alpha_t}}}{\sqrt{\bar{\alpha_t}}}\epsilon\right]\right]\right|
\label{eq:theorem}
\end{equation}

As argued in section 4 of the main paper, this results shows that an ideal initial conditions minimizes the semantic difference $|\mathcal{E}(x_{init}^v)-\mathcal{E}(x_{latent}^v)|$, which motivates us to use MANO based hand model for a low-score initialization. 

\subsection{Derivation of CHS loss weighing term}

In this section, we provide derivation for the CHS loss weighing term defined in Eq. 6 of the main paper.

\begin{equation}
\lambda_t^{chs} = \lambda_{max}^{chs}\left[\frac{t-t_{min}}{t_{max}-t_{min}}\right] + \lambda_{min}^{chs}\left[\frac{t_{max}-t}{t_{max}-t_{min}}\right]
\label{eq:lambdaChs}
\end{equation}

We first define $\lambda_t^{chs}$ annealing as a function of optimization iterations as follows:

\begin{equation}
\lambda_t^{chs} = \lambda_{max}^{chs} - (\lambda_{max}^{chs} - \lambda_{min}^{chs})\sqrt{\frac{i}{i_{max}}}
\label{eq:lambda}
\end{equation}

\noindent where, $i$ and $i_{max}$ denote the current and maximum optimization iterations respectively. This is similar in form to that of square-root time annealing proposed by ~\cite{hifa} as follows:

\begin{equation}
\begin{split}
& t = t_{max} - (t_{max} - t_{min})\sqrt{\frac{i}{i_{max}}} \\
& \Rightarrow \sqrt{\frac{i}{i_{max}}} = \frac{t_{max} - t}{t_{max} - t_{min}}
\label{eq:imax}
\end{split}
\end{equation}
 
\noindent Substituting Eq. \ref{eq:imax} into Eq. \ref{eq:lambda} we get,

\small
\begin{equation}
\begin{split}
\lambda_t^{chs} & = \lambda_{max}^{chs} - (\lambda_{max}^{chs} - \lambda_{min}^{chs})\frac{t_{max} - t}{t_{max} - t_{min}} \\
                           & = \lambda_{max}^{chs}\left[1 - \frac{t_{max} - t}{t_{max} - t_{min}}\right] + \lambda_{min}^{chs}\left[\frac{t_{max} - t}{t_{max} - t_{min}}\right] \\
                           & = \lambda_{max}^{chs}\left[\frac{t-t_{min}}{t_{max}-t_{min}}\right] + \lambda_{min}^{chs}\left[\frac{t_{max}-t}{t_{max}-t_{min}}\right]
\end{split}
\end{equation}
\normalsize

As explained in the main paper, we use this equation to adjust the weight of CHS loss function so that more weights is given at higher noise-timesteps so that the geometry does not degrade too much. We empirically choose $\lambda_{max}^{chs}$, $\lambda_{min}^{chs}$, $t_{max}$ and $t_{min}$ as 15000, 1000, 600 and 300 respectively.

\section{A.2 Additional Implementation Details}

\subsection{Hand Shape Initialization}

\noindent\textbf{Obtaining hand silhouette groundtruth: } As explained in the main paper, we use hand silhouette mask obtained from MANO mesh to initialize the NeRFs in stage 1. To this end, we first obtain hand mesh in diverse articulations using the code provided by MANO ~\cite{mano}. The hand mesh is then placed in a virtual 3D environment using open3D library ~\cite{open3d}. We place virtual cameras, corresponding to the sampled viewpoints in the same environment. Next, we obtain the depth map of the hand mesh, as observed by the virtual cameras using open3D APIs. Finally, we convert the depth map into a binary map to use as the hand silhouette ground truth from the required viewpoint.


\subsection{Skeleton based SDS}

We use ControlNet v1.1 model ~\cite{controlnet}, conditioned on OpenPose skeleton ~\cite{openpose} for SDS based optimization. We obtain hand skeleton from a given view point using the codes provided by MANO ~\cite{mano}. However, this skeleton excludes the fingertips. Hence, we add fingertip keypoints using vertex information from MANO hand mesh, resulting in a 3D hand skeleton of dimensionality ($21\times3$). Next, we transform this skeleton into OpenPose format and project it onto the 2D view space of the virtual camera, to generate the control input. 

\begin{figure*}[tb]
  \centering
  \includegraphics[width=\textwidth]{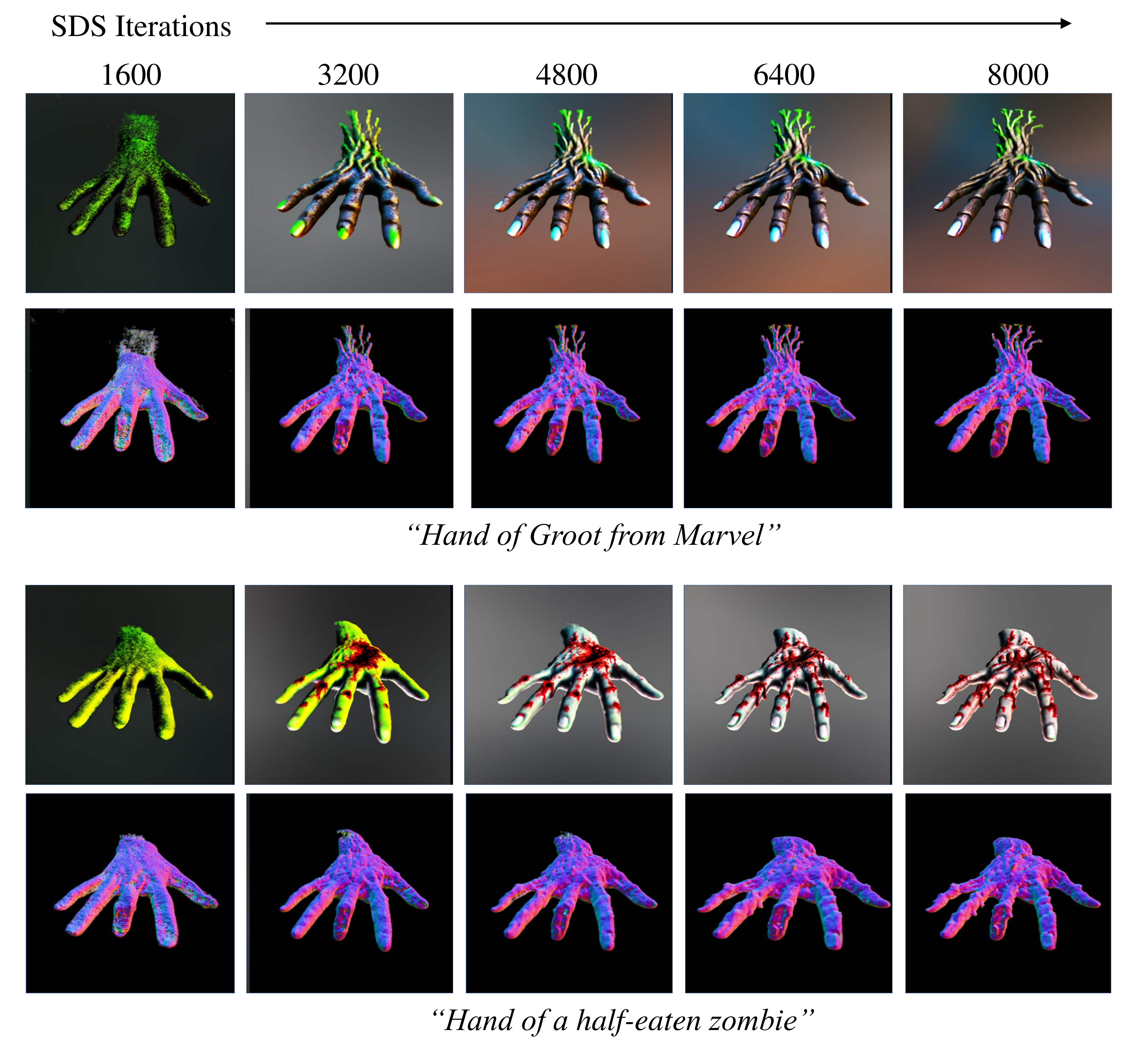}
  \caption{During the earlier iterations, the SDS optimizes geometry more and in the later iterations, texture is optimized more. It can be seen that the surface maps do not change much after 3200 iterations of SDS. This observation encourages us to anneal $\lambda_{t}^{chs}$ in Corrective Hand Shape guidance loss.}
  \label{fig:abl}
\end{figure*}

\section{A.3 Additional Studies}

In section 5.2 of the main paper, we justified CHS annealing by the observation that SDS tends to perform more geometric updates at higher noise $t$ (lower iteration) and more texture updates at lower $t$ (higher iteration). In Fig. \ref{fig:abl} we provide two examples on this observation. It can be seen that the geometry of the 3D model is optimized more in the earlier iterations and the texture in the later iterations. This empirically justifies the proposed CHS loss annealing, wherein we provide higher weightage to MANO  prior in the earlier iterations to ensure that the geometry optimization is stabilized.

\section{A.4 Additional Quantitative Analysis}

In this section, we provide analysis of both mean and standard deviation of the proposed method compared to state-of-the-art in Table. 1. While methods such as OHTA ~\cite{ohta} and Fantasia3D ~\cite{fantasia} generates results with lower standard deviation for CLIP L14, their mean scores are much lower compared to our method. Further, state of the art method CFD ~\cite{cfd} achieves the second best mean scores on all the metrics, but they have a high standard deviation in their results. It can be seen our method (HandDreamer) achieves the best mean value for all the 3 metrics with a low standard deviation. This shows that our method generates the best results consistently over several prompts. 

\section{A.5 Additional Results}

\begin{figure*}[tb]
  \centering
  \includegraphics[width=\textwidth]{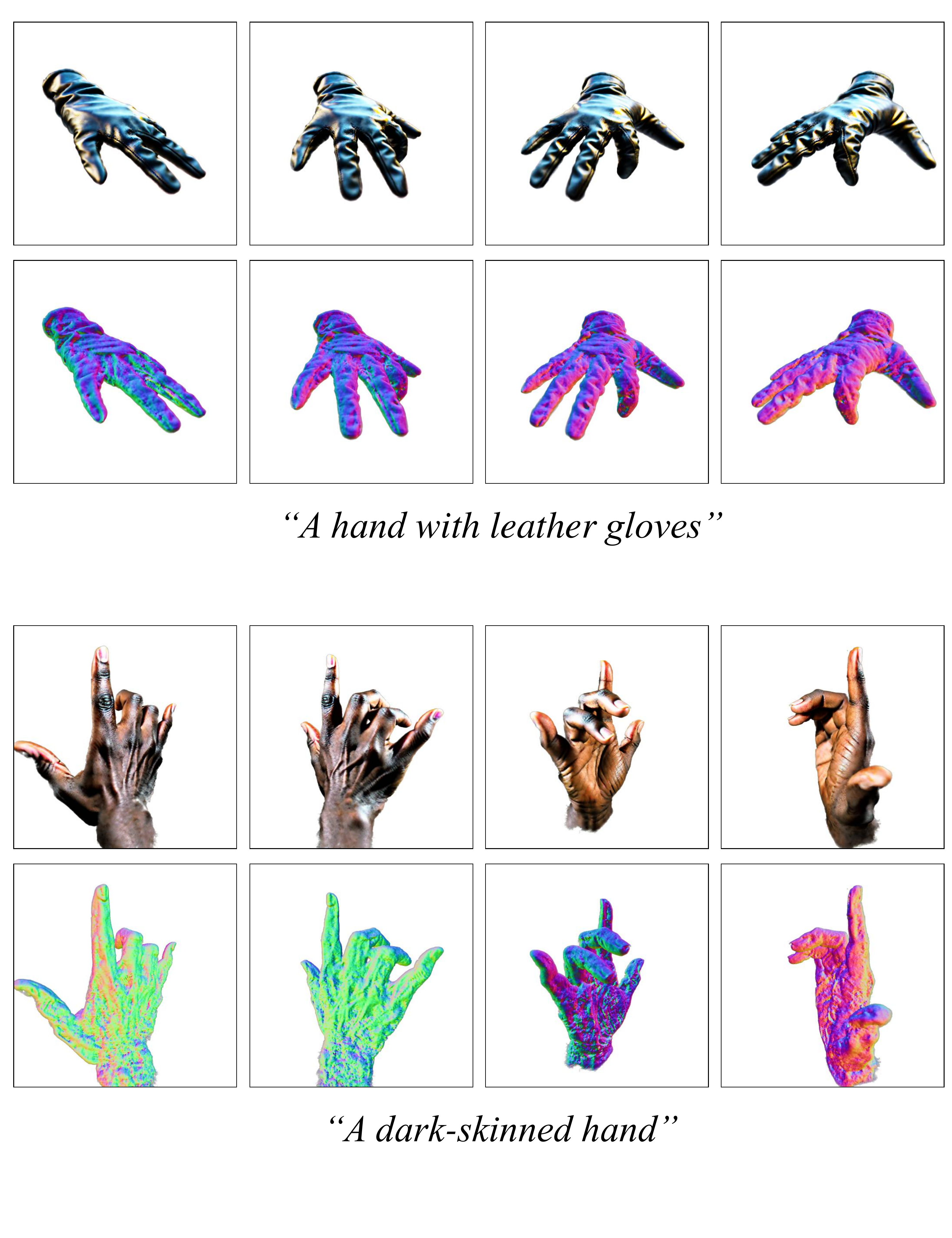}
  \caption{Results from the proposed HandDreamer method. Top row: Rendered images. Bottom Row: Surface maps}
  \label{fig:results1}
\end{figure*}

\begin{figure*}[tb]
  \centering
  \includegraphics[width=\textwidth]{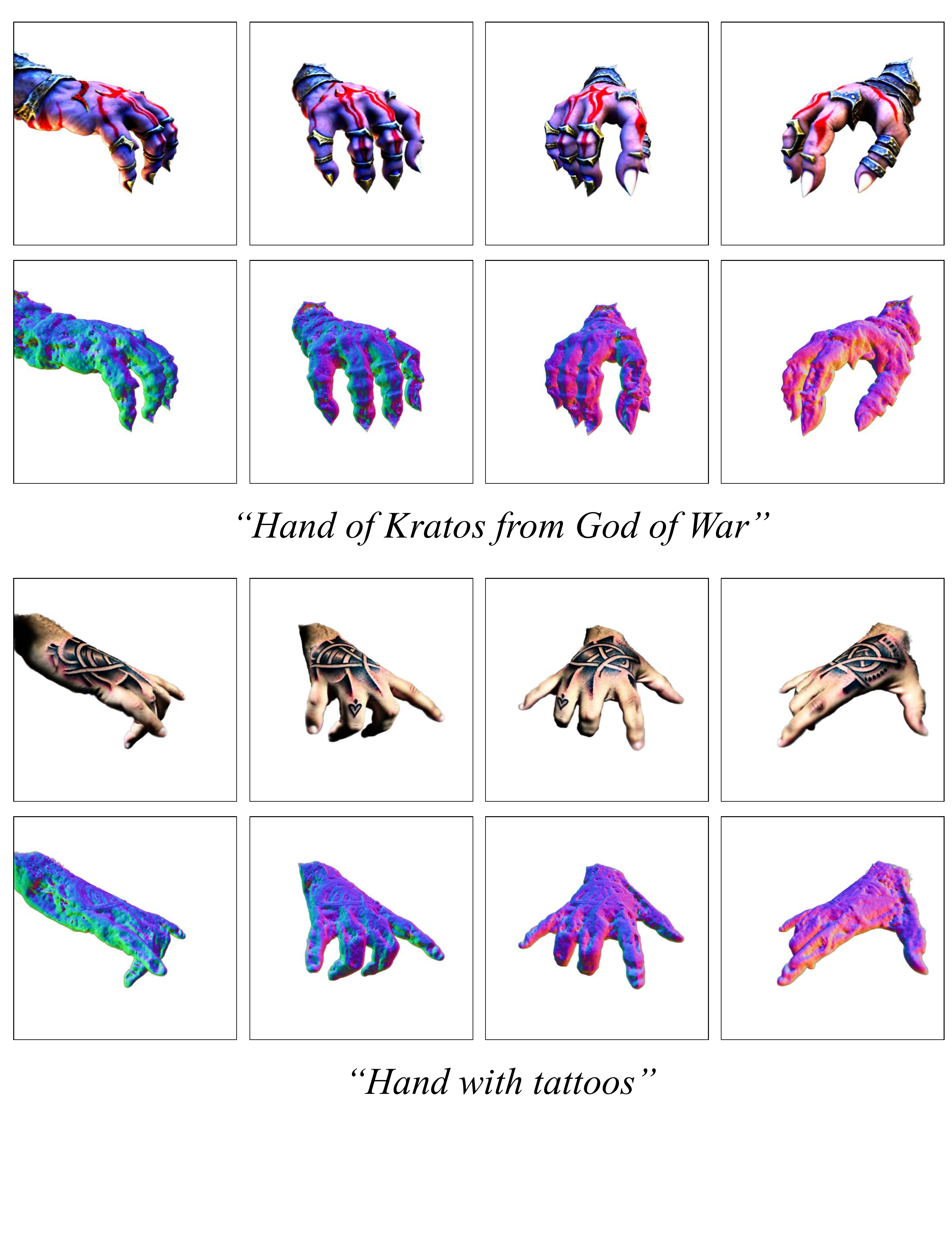}
  \caption{Results from the proposed HandDreamer method. Top row: Rendered images. Bottom Row: Surface maps}
  \label{fig:results4}
\end{figure*}

\begin{figure*}[tb]
  \centering
  \includegraphics[width=\textwidth]{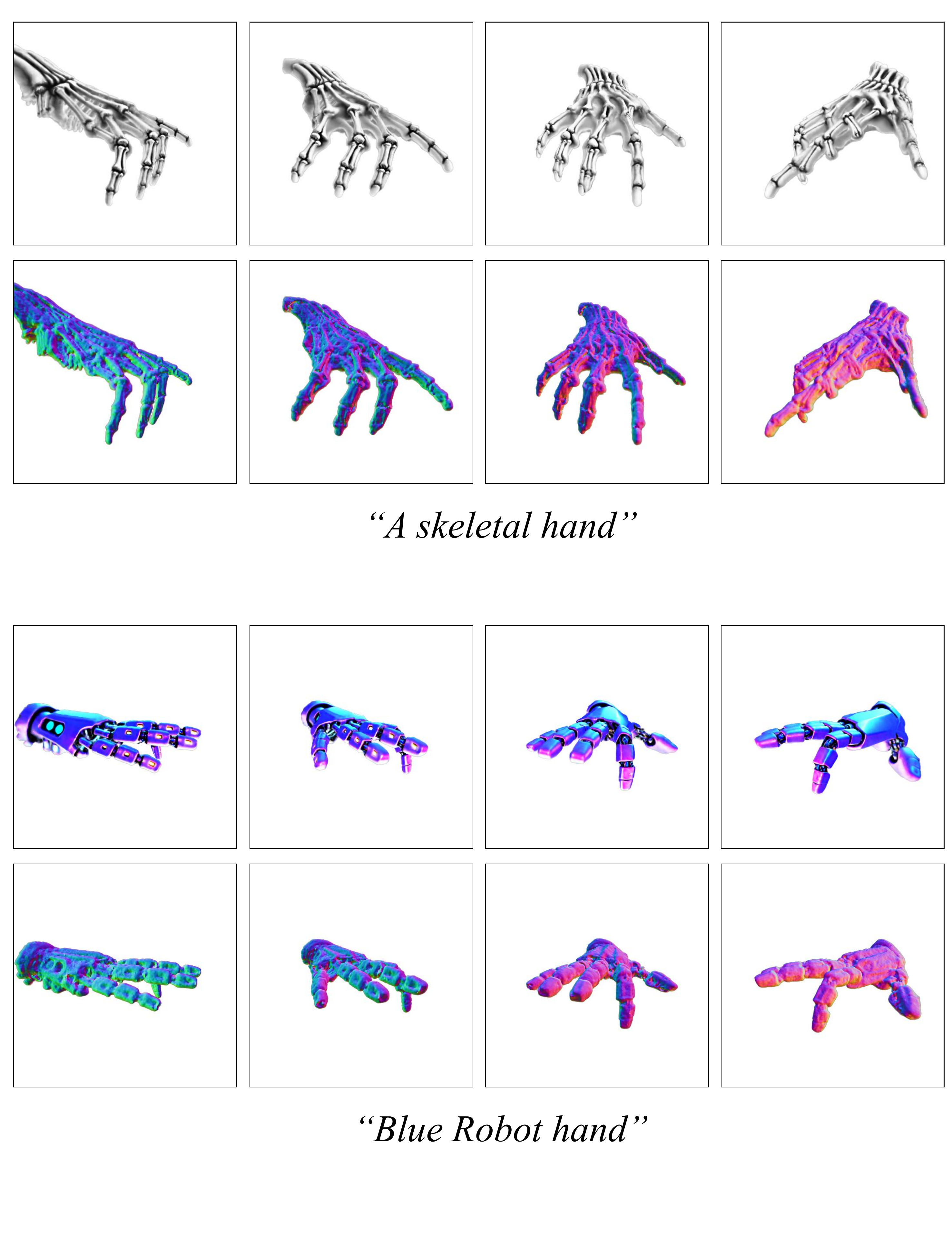}
  \caption{Results from the proposed HandDreamer method. Top row: Rendered images. Bottom Row: Surface maps}
  \label{fig:results5}
\end{figure*}

\begin{figure*}[tb]
  \centering
  \includegraphics[width=\textwidth]{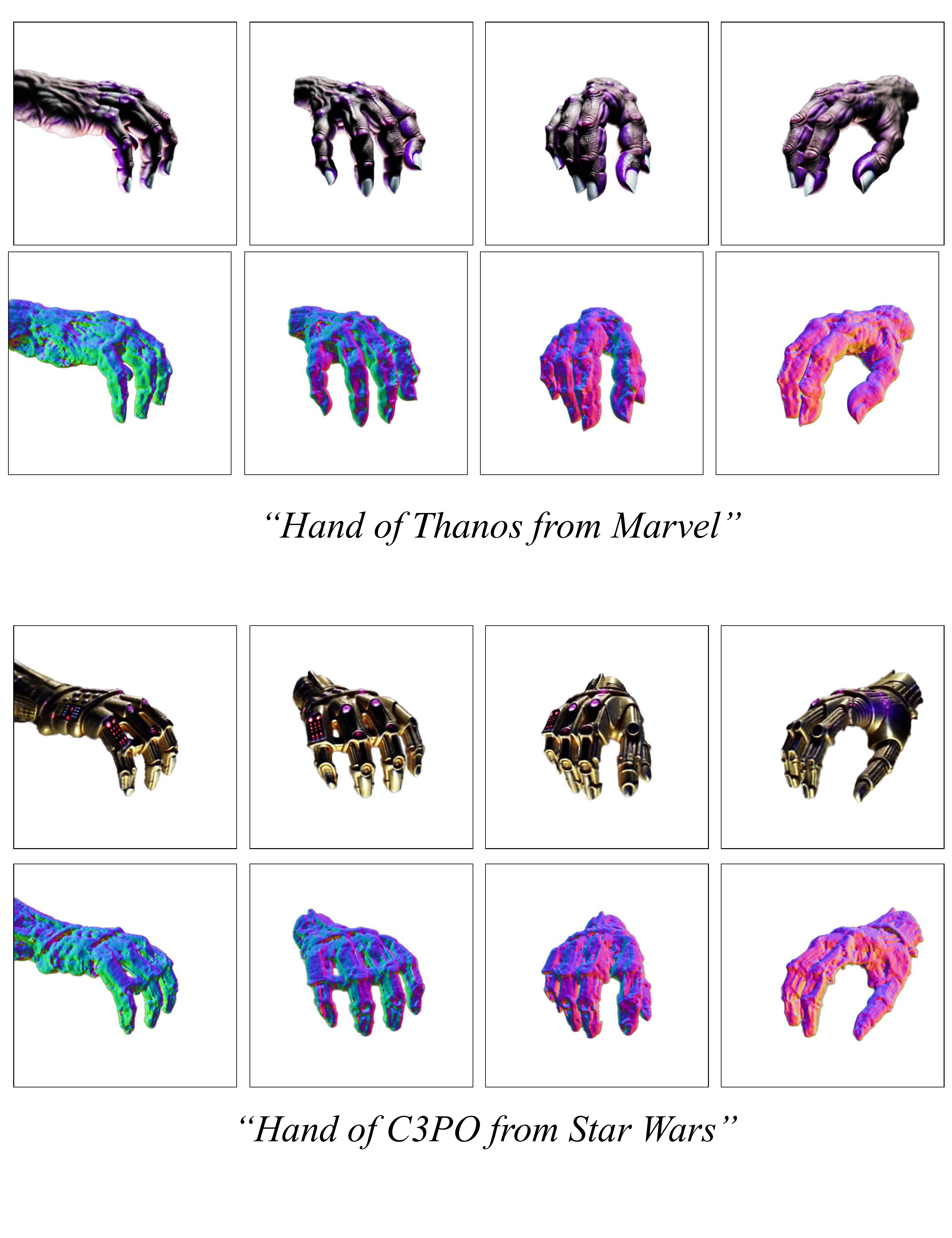}
  \caption{Results from the proposed HandDreamer method. Top row: Rendered images. Bottom Row: Surface maps}
  \label{fig:results6}
\end{figure*}

We provide additional results from our method for multiple viewpoints for several prompts in Figures \ref{fig:results1} to \ref{fig:results6}. It can be seen that our method generates high fidelity 3D models for a variety of text prompts. We have also provided videos in the multimedia supplementary.

\begin{figure*}[tb]
  \centering
  \includegraphics[width=\textwidth]{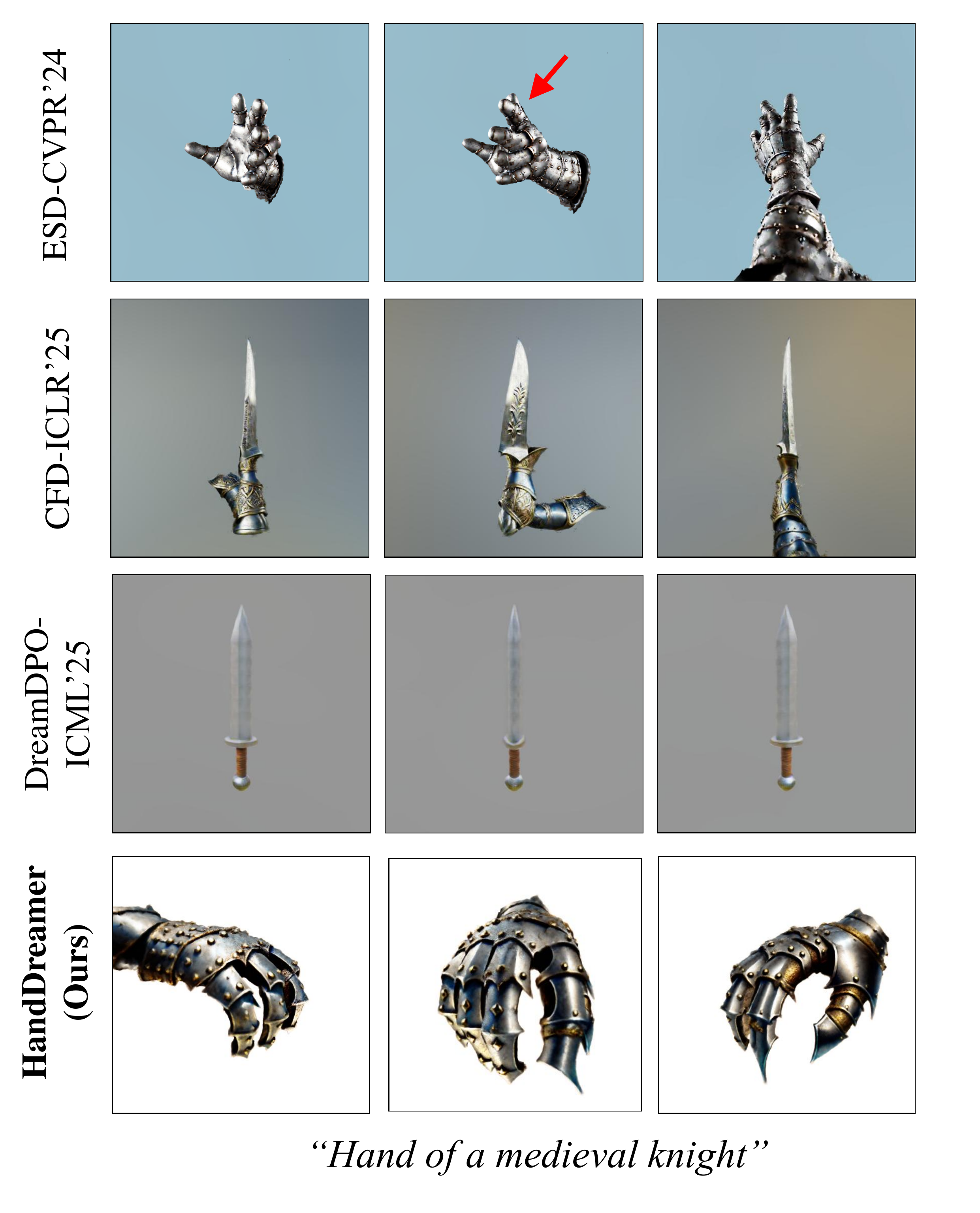}
  \caption{Comparisons against state-of-the-art methods: ESD'24 ~\cite{modeCollapse}, CFD'25 ~\cite{cfd}, DreamDPO ~\cite{dreamdpo}. ESD generates Janus effects (red arrows) while CFD and dreamDPO fail to generate a hand}
  \label{fig:results7}
\end{figure*}

\begin{figure*}[tb]
  \centering
  \includegraphics[width=\textwidth]{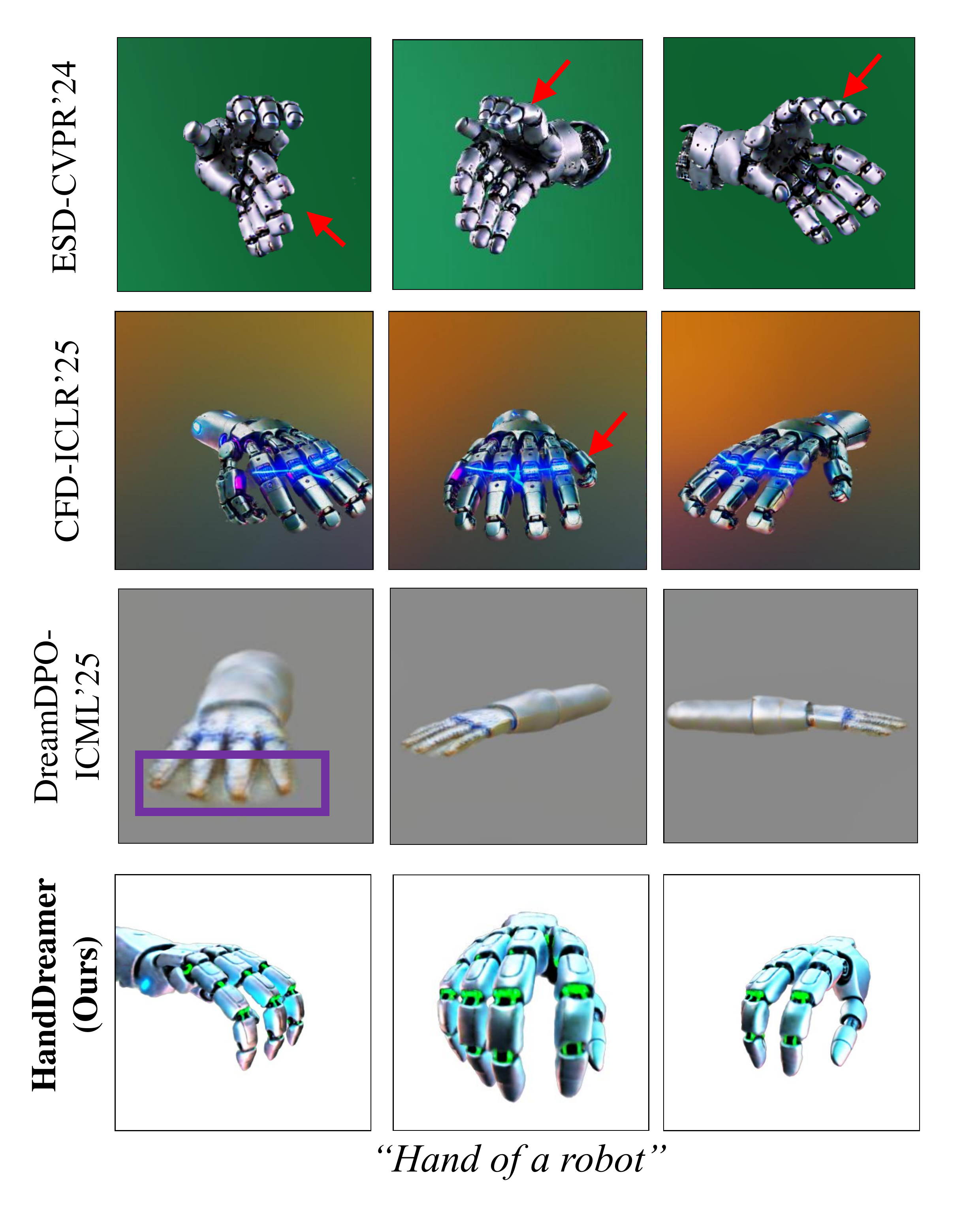}
  \caption{Comparisons against state-of-the-art methods: ESD'24 ~\cite{modeCollapse}, CFD'25 ~\cite{cfd}, DreamDPO ~\cite{dreamdpo}. Janus effects (ESD) and extra finger artifacts (CFD) shown in red arrows. Missing fingers (dreamDPO) denoted in violet box. }
  \label{fig:results8}
\end{figure*}

\begin{figure*}[tb]
  \centering
  \includegraphics[width=\textwidth]{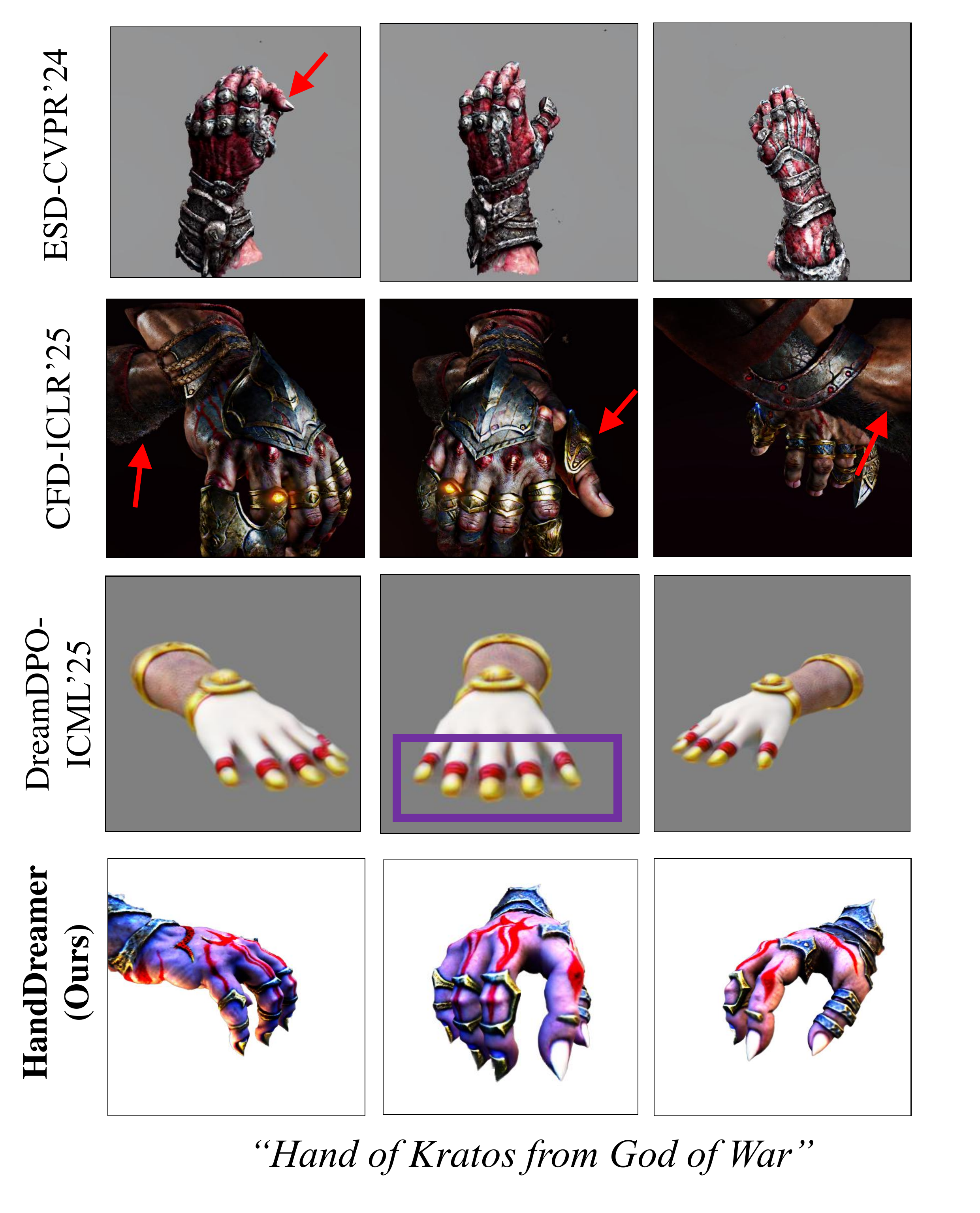}
  \caption{Comparisons against state-of-the-art methods: ESD'24 ~\cite{modeCollapse}, CFD'25 ~\cite{cfd}, DreamDPO ~\cite{dreamdpo}. ESD generates Janus artifacts (red arrows) and CFD generates extra fingers and arms (red arrows). DreamDPO generates low-fidelity hands with same length for all fingers (violet box)}
  \label{fig:results9}
\end{figure*}

\begin{figure*}[tb]
  \centering
  \includegraphics[width=\textwidth]{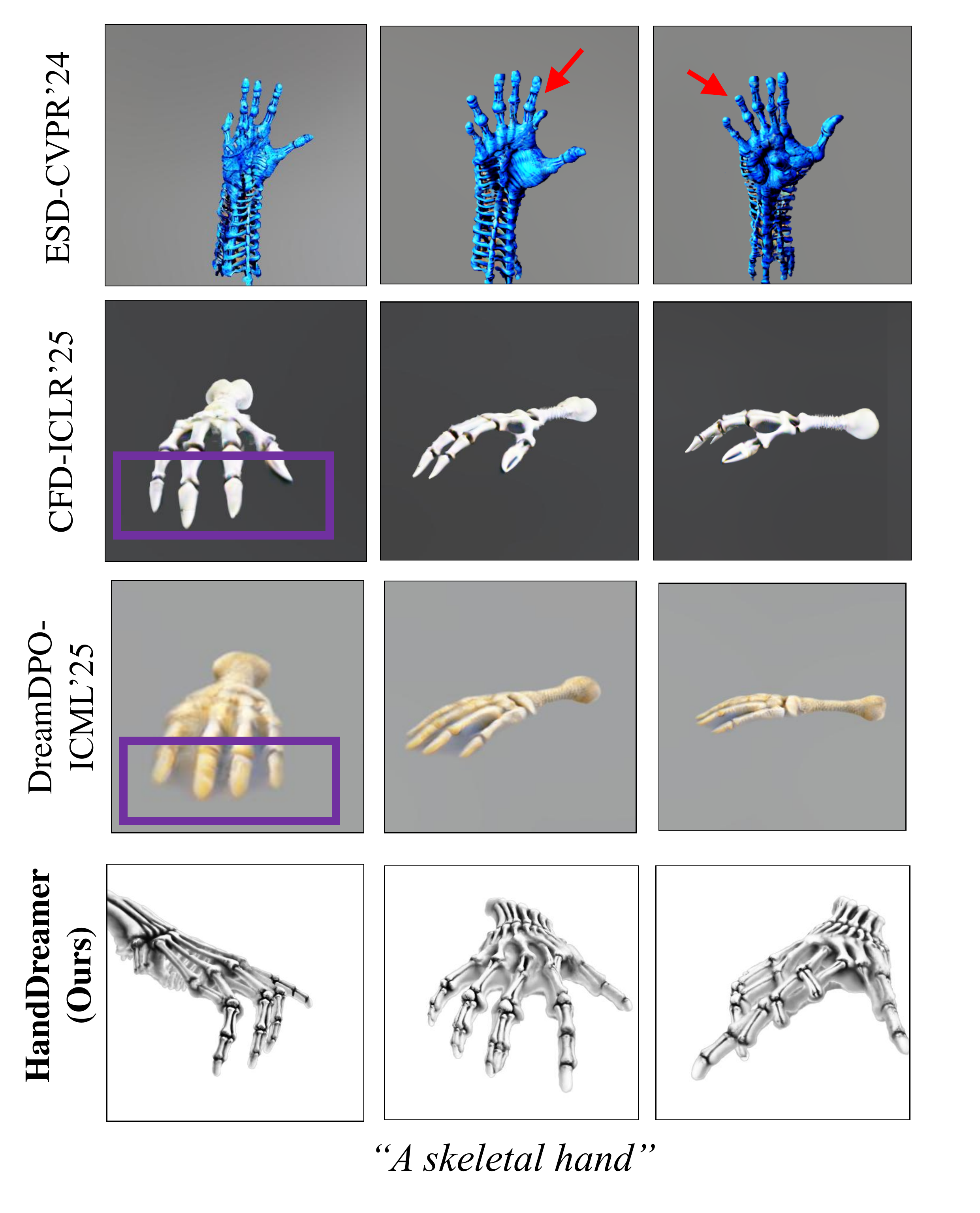}
  \caption{Comparisons against state-of-the-art methods: ESD'24 ~\cite{modeCollapse}, CFD'25 ~\cite{cfd}, DreamDPO ~\cite{dreamdpo}. Janus effects shown in red arrows (ESD). Missing fingers denoted in violet box (CFD, dreamDPO).}
  \label{fig:results10}
\end{figure*}

\begin{figure*}[tb]
  \centering
  \includegraphics[width=\textwidth]{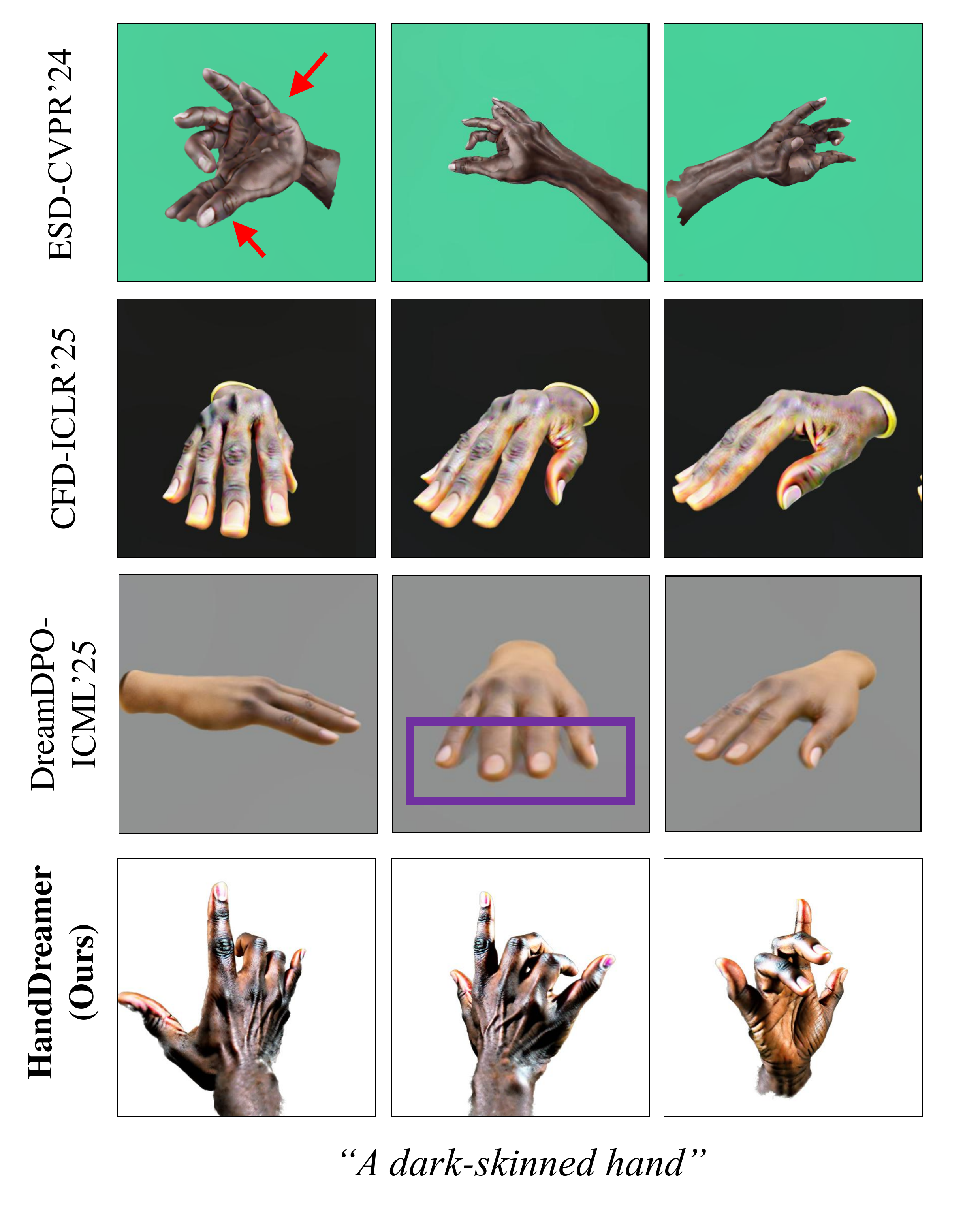}
  \caption{Comparisons against state-of-the-art methods: ESD'24 ~\cite{modeCollapse}, CFD'25 ~\cite{cfd}, DreamDPO ~\cite{dreamdpo}. Janus effects shown in red arrows (ESD). Missing fingers denoted in violet box (dreamDPO).}
  \label{fig:results11}
\end{figure*}

We also provide additional comparative studies against state-of-the art methods ESD'24 ~\cite{modeCollapse}, CFD'25 ~\cite{cfd} and dreamDPO'25 ~\cite{dreamdpo} in Figs. \ref{fig:results7} to \ref{fig:results11}. It can be seen that while ESD generates Janus artifacts with protruding fingers, CFD and dreamDPO generates results with erroneous number of fingers and lower details. On the other hand, our method is able to generate high-fidelity and geometrically accurate 3D hand model outputs. 

\setlength{\tabcolsep}{8pt}
\begin{table*}[t]
\begin{center}

\begin{tabular}{cccc}
\hline
	Method & CLIP L14 $\uparrow$ & FID $\downarrow$ & HPSv2 $\uparrow$\\
\hline  \\[-1.5ex]
DreamFusion'22 ~\cite{dreamfusion}&  25.12 \textpm 2.41 & 344.19 \textpm 45.54 & 0.187 \textpm 0.035 \\
LatentNerf'23 ~\cite{latentnerf}& 24.34 \textpm 3.61 & 316.42 \textpm 34.02 & 0.189 \textpm 0.026 \\
Fantasia3D'23 ~\cite{fantasia} & 20.93 \textpm 1.21 & 329.31 \textpm 62.28 & 0.198 \textpm 0.013\\
DreamWaltz'23 ~\cite{dreamwaltz}& 23.96 \textpm 3.08 & 265.11 \textpm 37.32 & 0.222 \textpm 0.021 \\
DreamAvatar'24 ~\cite{dreamavatar}& 20.02 \textpm 2.12& 329.85 \textpm 50.02 & 0.215 \textpm 0.025\\
HumanNorm'24 ~\cite{humannorm}& 23.01 \textpm 2.56& 327.42 \textpm 34.32 & 0.177 \textpm 0.021\\
SDI'24 ~\cite{sdi}& 26.32 \textpm 3.12 & 297.12 \textpm 35.32 & 0.192 \textpm 0.013 \\
OHTA'24 ~\cite{ohta}& 22.59 \textpm 0.93 & 467.51 \textpm 21.01 & 0.181 \textpm 0.017 \\
CFD'25 ~\cite{cfd}& \underline{26.62 \textpm 5.12} & \underline{262.83 \textpm 44.32} & \underline{0.223 \textpm 0.021} \\

\textbf{HandDreamer (Ours)}  & \textbf{28.63 \textpm 1.49}& \textbf{254.62 \textpm 34.12} & \textbf{0.241 \textpm 0.012} \\
\hline
\end{tabular}
\caption{Quantitative comparisons. Our method outperforms the other methods on all the metrics while generating results in low standard deviation.}
\end{center}
\label{tab:sota1}
\end{table*}

\begin{figure*}[tb]
  \centering
  \includegraphics[width=\textwidth]{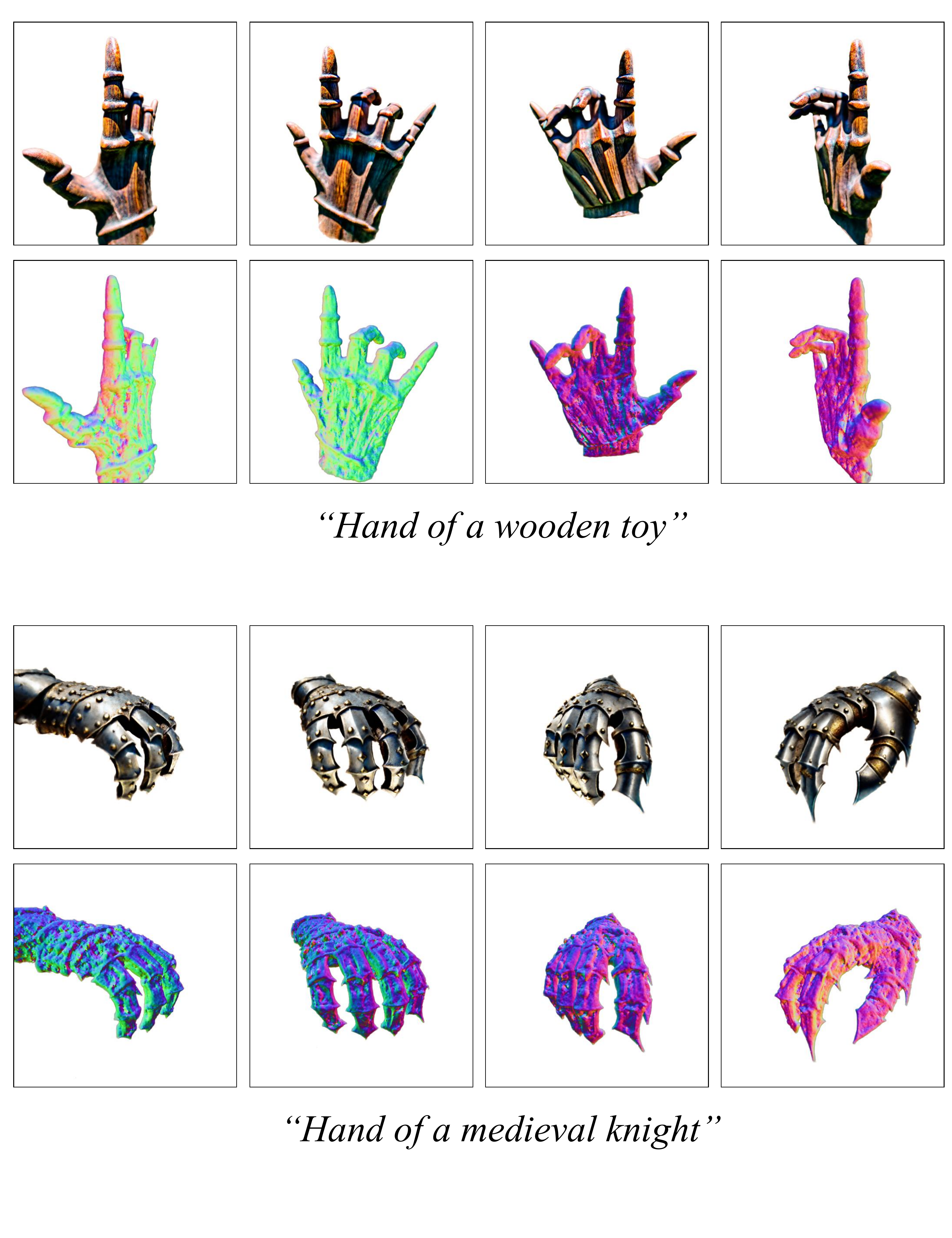}
  \caption{Results from the proposed HandDreamer method. Top row: Rendered images. Bottom Row: Surface maps}
  \label{fig:results3}
\end{figure*}

\begin{figure*}[tb]
  \centering
  \includegraphics[width=\textwidth]{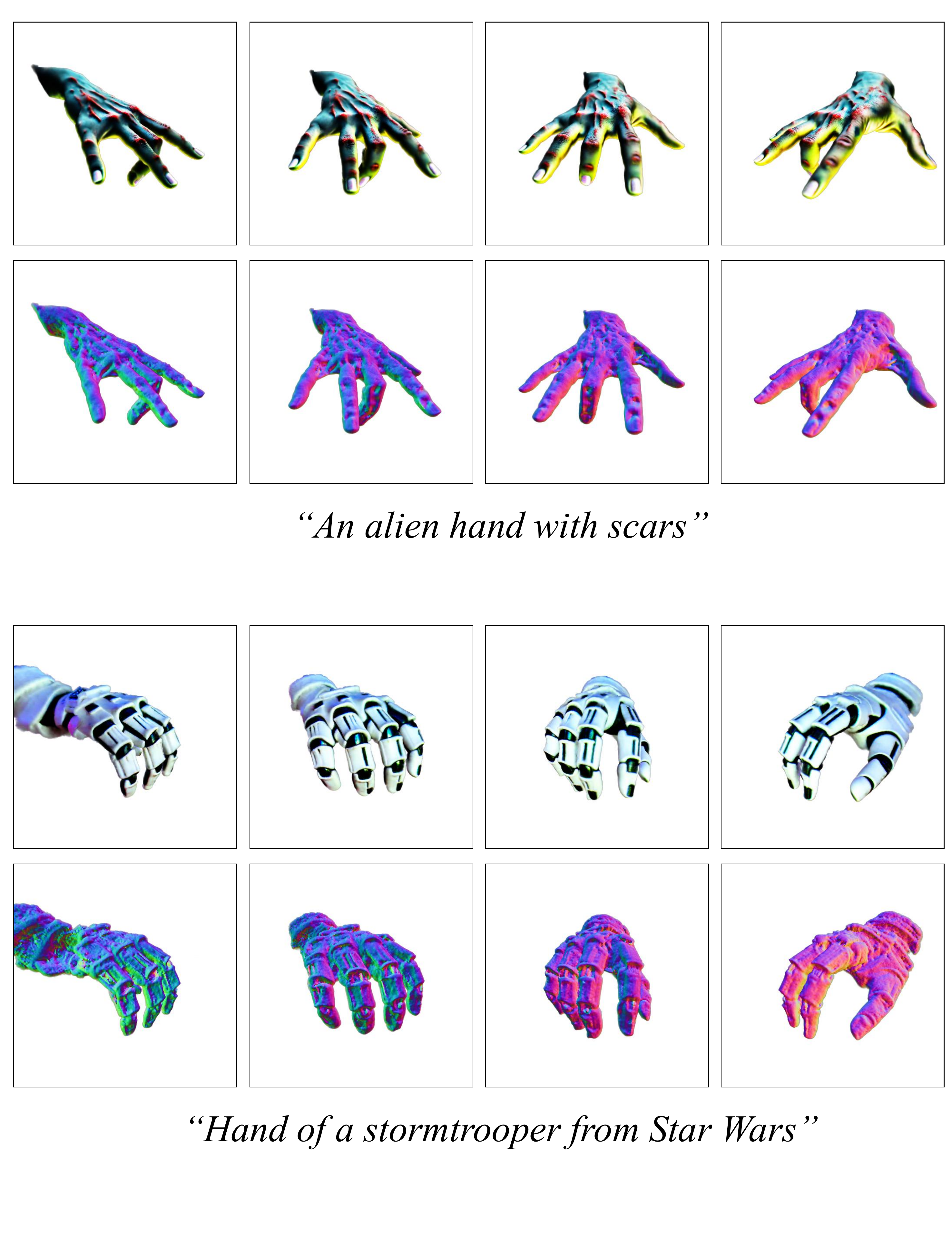}
  \caption{Results from the proposed HandDreamer method. Top row: Rendered images. Bottom Row: Surface maps}
  \label{fig:results3}
\end{figure*}

\end{document}